\pdfoutput=1

\documentclass[11pt]{article}

\usepackage[final]{acl}

\usepackage{times}
\usepackage{latexsym}
\usepackage{tabularx}
\usepackage[T1]{fontenc}

\usepackage[utf8]{inputenc}

\usepackage{microtype}

\usepackage{inconsolata}

\usepackage{graphicx}
\usepackage{pifont}
\usepackage{xcolor}

\usepackage{algorithm}
\usepackage[utf8]{inputenc} %
\usepackage[T1]{fontenc}    %
\usepackage{hyperref}       %
\usepackage{url}   
\usepackage{array}
\usepackage{booktabs}       %
\usepackage{amsfonts}       %
\usepackage{nicefrac}       %
\usepackage{microtype}      %
\usepackage{xcolor}         %
\usepackage{listings}
\usepackage{float}
\usepackage{amsmath}
\usepackage{algpseudocode}
\usepackage{multirow}
\definecolor{codegreen}{rgb}{0,0.6,0}
\definecolor{codegray}{rgb}{0.5,0.5,0.5}
\definecolor{codepurple}{rgb}{0.58,0,0.82}
\definecolor{backcolour}{rgb}{0.95,0.95,0.92}

\lstdefinestyle{mystyle}{
    backgroundcolor=\color{backcolour},   
    commentstyle=\color{codegreen},
    keywordstyle=\color{magenta},
    numberstyle=\tiny\color{codegray},
    stringstyle=\color{codepurple},
    basicstyle=\ttfamily\tiny,
    breakatwhitespace=false,         
    breaklines=true,                 
    captionpos=b,                    
    keepspaces=true,                 
    numbersep=5pt,                  
    showspaces=false,                
    showstringspaces=false,
    showtabs=false,                  
    tabsize=2
}

\usepackage{titlesec}
\titlespacing*{\paragraph}{0ex}{0.5ex}{1ex}
\titlespacing*{\subsubsection}{0ex}{0.5ex}{1ex}
\titleformat*{\subsubsection}{\itshape}

\DeclareMathOperator*{\argmax}{arg\,max}

\title{Optimizing Instructions and Demonstrations \\for Multi-Stage Language Model Programs}

\author{
 \textbf{Krista Opsahl-Ong\textsuperscript{1}\thanks{Equal contribution.}},\hspace{1mm}
 \textbf{Michael J Ryan\textsuperscript{1}\footnotemark[1]},\hspace{1mm}
 \textbf{Josh Purtell\textsuperscript{2}},\hspace{1mm}
\\
 \textbf{David Broman\textsuperscript{3}},\hspace{1mm}
 \textbf{Christopher Potts\textsuperscript{1}},\hspace{1mm}
 \textbf{Matei Zaharia\textsuperscript{4}},\hspace{1mm}
 \textbf{Omar Khattab\textsuperscript{1}}
\\
\\
 \textsuperscript{1}Stanford University,
 \textsuperscript{2}Basis,
 \textsuperscript{3}KTH Royal Institute of Technology
 \textsuperscript{4}UC Berkeley
}

\begin{document}
\maketitle
\begin{abstract}
Language Model Programs, i.e.\ sophisticated pipelines of modular language model (LM) calls, are increasingly advancing NLP tasks. However, building these pipelines requires crafting prompts that are jointly effective for all modules. We study prompt optimization for LM programs, i.e.\ how to update these prompts to maximize a downstream metric without access to module-level labels or gradients. To make this tractable, we factorize our problem into optimizing the free-form instructions and few-shot demonstrations of every module and introduce several strategies to craft task-grounded instructions and navigate credit assignment across modules. Our strategies include (i) program-and-data-aware techniques for proposing effective instructions, (ii) a stochastic mini-batch evaluation function for learning a surrogate model of our objective, and (iii) a meta-optimization procedure in which we refine how LMs construct proposals over time. Using these insights we develop MIPRO, a novel optimizer that outperforms baselines on five of seven diverse LM programs using a best-in-class open-source model (Llama3-8B), by as much as 13\% accuracy. We have released our new optimizers and benchmark in DSPy at \url{http://dspy.ai}.

\end{abstract}

\section{Introduction}
\label{sec:intr}

\begin{figure}[ht]
    \centering
  \includegraphics[width=0.35\textwidth]{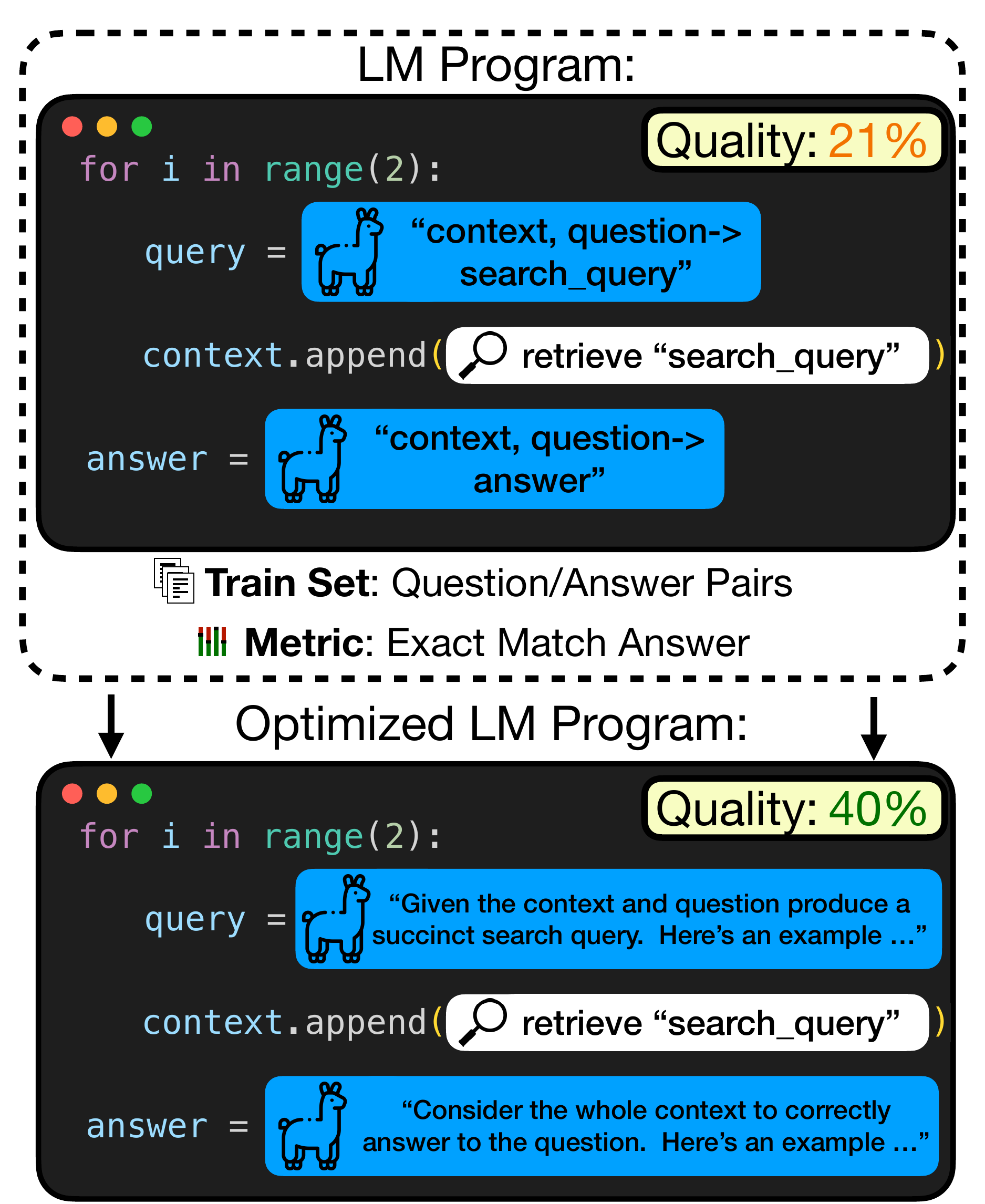}
  \caption{An example of the optimization problem we explore, shown for a multi-hop retrieval LM program. Given some question--answer pairs and a metric, the optimizer proposes new instructions and bootstraps new demonstrations (not pictured) for each stage.
  }
  \label{fig:motivation}
  \vspace{-5mm}
\end{figure}

Solving complex tasks with Language Models (LMs) often requires applying sophisticated prompting techniques~\citep{wei2022chain,chen2022program} and chaining them together into multi-stage pipelines ~\citep{wu2022ai,dohan2022language,khattab2022demonstrate,beurer2023prompting,yao2023react,schlag2023large}. Such \textit{Language Model (LM) Programs} continue to advance NLP tasks~\citep{pourreza2023din,khattab2024dspy,ridnik2024code} through systematic composition \citep{khattab2021baleen,creswell2022faithful, pan2024autonomous} and tool use~\citep{qin2023toolllm}.  However, LM programs today are commonly designed via ``prompt engineering'': crafting prompts via manual trial and error to coerce a specific LM to operate each step in a specific pipeline. 
Recent work such as APE \citep{zhou2023large}, OPRO \citep{yang2024large}, and EvoPrompt \citep{guo2024connecting} %
presents \textit{prompt optimizers}, i.e.,\ algorithms that search over strings to identify high-performing prompts. Unfortunately, the majority of this work does not directly apply to multi-stage LM programs in which we lack gold labels or evaluation metrics for the individual LM calls.
\citet{khattab2024dspy} study how to express \textit{arbitrary LM pipelines} such that the instructions, demonstrations (input/output examples), and LM weights of each LM call are treated as parameters that can be optimized toward any metric. While the authors present optimizers that can create demonstrations of multi-stage pipelines and use them to optimize prompts, weights, or even both together \cite{soylu2024finetuningpromptoptimizationgreat}, their proposed optimizers cannot tune instructions for multi-prompt pipelines.

We seek to efficiently optimize prompts in arbitrary LM programs, especially those with multiple stages (Figure~\ref{fig:motivation}) and explore approaches that hold under weak assumptions, consistent with the abstractions from the DSPy programming model~\citep{khattab2024dspy}. In particular, we assume no access to LM weights, log-probabilites, or handwritten metrics or labels for intermediate stages in a chain of LM calls. We require only the LM program itself, a metric to optimize, and a training set of inputs (and, depending on the metric, final outputs).

We formally define the problem of prompt optimization for LM programs and outline the design space by identifying two key challenges. First, the \emph{proposal challenge}: the space of possible prompts is intractably large, and this is exacerbated as the number of modules increase.  Proposing a few high-quality instructions is thus essential. Second, the \emph{credit assignment challenge}: our problem requires jointly optimizing over many distinct variables that parameterize the prompts of all modules. To allocate search effort, we must infer the impact of our configurations for each variable effectively.

We define several strategies to tackle each of these challenges and systematically explore the tradeoffs they present. We find that optimizing bootstrapped few-shot examples is often essential for realizing the greatest performance gains, but that optimizing instructions becomes more essential for tasks with conditional rules. We also find that optimizing both instructions and few-shot examples together generally leads to the best results.

We make three contributions. First, we present a formalization of the problem of optimizing language model programs (\S \ref{sec:problem-statement}) and propose an algorithm design space with three strategies to address the challenge of prompt proposal and three strategies to resolve the issue of credit assignment (\S \ref{sec:design}). Second, we release a benchmark suite for LM program optimizers spanning seven tasks (\S\ref{sec:experiments}). Third, we construct and evaluate a rich subset of possible algorithms for prompt optimization (\S \ref{sec:optimizers}).  Highlighted amongst these algorithms is MIPRO (\underline{M}ulti-prompt \underline{I}nstruction \underline{PR}oposal \underline{O}ptimizer) which outperforms  baseline optimizers on five of seven tasks in our benchmark, by as much as 13\% accuracy improvement.  Using our algorithms, we derive five key lessons for practitioners looking to optimize LM programs (\S \ref{sec:results_and_discussion}).

\section{Problem Statement}
\label{sec:problem-statement}

\begin{algorithm}
\caption{Optimize $\Phi$ with optimizer $M$} %
\small
\begin{algorithmic}[1]
  \State \textbf{Input:}
  Optimizer $M$,
  Initial Program $\Phi$,
  Metric $\mu$
  \State \textbf{Input:} Max Iterations $I$,
  Training Data $\mathcal{D}$
  \State \textbf{Input:} Minibatch size $B$,
  Proposer Hyperparameters $\theta$
\State \textbf{Output:} Optimized version of $\Phi$
\State
\State %
$M$.Initialize($\mathcal{D}$, $\theta$) \Comment{Initialize optimizer using the data}
\For{$k \gets 1 \textbf{ to } I$}
    \State $(\mathbf{V} \mapsto S_{k}) \gets$ $M$.Propose($\theta$) \Comment{Generate proposal}
    \State $\mathcal{D}_{k} \gets \{(x_j,x'_j) \sim \mathcal{D}\}_{j=1}^B$ \Comment{Sample size-$B$ batch} 
    \State $\sigma \gets \frac{1}{B}\sum_{(x, x') \in \mathcal{D}_{k}}\mu(\Phi_{\mathbf{V} \mapsto S_{k}}(x), x')$ \Comment{Validate updated program}
    \State %
    $M$.Update($\mathbf{V} \mapsto S_{k}$, $\sigma$) \Comment{Update optimizer based on the observed validation score}
\EndFor
\State $(\mathbf{V} \mapsto S_{k})$ $\gets$ $M$.ExtractOptimizedSets()
\State \textbf{return} $\Phi_{\mathbf{V} \mapsto S}$
\end{algorithmic}
\label{alg:general}
\end{algorithm}

Consider an LM program $\Phi$ consisting of $m$ modules, each using some LM. Each module $i$ is defined by a prompt template $p_{i}$ that contains a set of variables (open slots) $\mathbf{v}$. For example, a prompt template for few-shot QA might have variables for instructions, demonstrations, and the target question. 

Let $\mathbf{V}$ be the set of all variables used by prompt templates for $\Phi$, and 
let $\mathbf{V} \mapsto S$ be a total assignment of variables $\mathbf{V}$ to strings $S$. We use $\Phi_{\mathbf{V} \mapsto S}$ to specify the program $\Phi$ run under such an assignment. Our high-level goal is to find a total assignment that optimizes $\Phi$'s performance 
with respect to metric $\mu$ on a trainset $\mathcal{D}$ that has inputs $X$ and optional metadata $X'$ (such as labels):
\begin{equation*}
  \Phi^* =
  \argmax_{\mathbf{V} \mapsto S}
  \, \frac{1}{|\mathcal{D}|} \sum_{(x, x') \in \mathcal{D}}
  \mu(\Phi_{\mathbf{V} \mapsto S}(x), x')
  \label{eq:prompt_components_new}
\end{equation*}
This is the problem faced by people designing LM programs. It is intractable, as (i) each string $s \in S$ can take on any value, (ii) the metric $\mu$ provides supervision only at the level of the entire task, so every variable in $\mathbf{V}$ is latent, and (iii) we assume no access to the gradients or embeddings of the LMs involved, which rules out many RL and prompt-tuning algorithms \citep{zhang2022tempera, li-liang-2021-prefix, shin-etal-2020-autoprompt}. In addition, (iv) system designers generally have small datasets $\mathcal{D}$ and (v) small budgets of LM calls for evaluating $\Phi$.

In many cases, we want to optimize just a subset of the variables used by $\Phi$. In the present work, for example, we assume each prompt~$p$ has a variable~$i$ over free-form instructions and a set of~$K$ variables $\{d_{i1}, \ldots, d_{ik}\}$ over demonstrations. In these settings, we assume that all the other variables for $\Phi$ are set to constant values.

To find approximate solutions to \eqref{eq:prompt_components_new}, we work within the general optimization framework defined by Algorithm~\ref{alg:general}.  This framework generalizes prior approaches such as OPRO \citep{yang2023large} and APE \citep{zhou2022large} to optimizing LM programs. The main parameters to this method are the optimizer $M$ and the unoptimized program $\Phi$. We assume that each optimizer has methods Initialize, Propose, Update, and ExtractOptimizedSets and that it has some internal state which informs proposals, and updates on calls to update.

\section{Designing LM Program Optimizers}
\label{sec:design}
Algorithm~\ref{alg:general} defines a general optimization framework for LM programs. We seek efficient instantiations of this algorithm in which we minimize the number of times the program $\Phi$ is invoked (Line~10) and, as a result, the number of times we must sample proposals (Line~8). To this end, we are especially interested in building LM program optimizers with strategies that handle the proposal and credit assignment challenges discussed in Section~\ref{sec:intr}. In this section, we present several novel or improved strategies for this. Section~\ref{sec:optimizers} then defines a few effective compositions of these strategies that allow us to empirically study their properties in practice.  We showcase how these components come together to make LM Program optimizers in Figure~\ref{fig:MIPRO_figure}.

\subsection{The Proposal Problem}
\label{sec:prompt_proposal}

To make the approximate optimization tractable, we must be able to efficiently sample candidate prompts that are well suited to the nature of our task, program, data, and metric. To do this, we leverage another LM as a `proposer' LM, and consider (i) bootstrapping few-shot examples that demonstrate how to conduct the task, (ii) collecting and summarizing important factors that could inform the construction of high-quality instructions, and/or (iii) meta-optimizing how the proposer LM is used to create high-performing instructions.

\textbf{Bootstrapping Demonstrations}
\citet{khattab2022demonstrate,khattab2024dspy} study a simple yet surprisingly effective rejection-sampling strategy for optimizing the prompts of LM programs. Inputs $x$ are sampled from the training set, and run through $\Phi(x)$ to generate input/output traces $\tau$ for each module in the program. If the output scores as measured by metric $\mu$ are successful, i.e.\ $\mu(\Phi(x), x') \geq \lambda$ for some threshold $\lambda$ and 
$(x, x') \in \mathcal{D}$, they treat all values in the trace as a potential labeled demonstrations (i.e. valid input/output examples) for the respective module in $\Phi$. 
Given these potential demonstrations, the optimization problem is reduced to selecting combinations of demonstrations (within and across modules) that serve as effective few-shot examples for prompting. \citet{khattab2024dspy} find this can often outperform hand-written demonstrations for multi-stage programs.

\textbf{Grounding} How can we guide our proposal LM to craft performant instructions for a given module? We hypothesize that providing the proposal LM with relevant context, such as properties of the data, the program, and examples of successful task completions, will allow it to create instructions better suited to the task. Hence, we build a zero-shot LM program for (i) characterizing patterns in the raw dataset $\mathcal{D}$, (ii) summarizing the program's control flow, (iii) bootstrapping program demonstrations, and (iv) collecting per-stage prompts that were previously evaluated with their evaluation scores on the train set.  We consider supplying each of these pieces as context to `ground' the LM proposing our instructions. Details on constructing the dataset and program summaries are included in Appendix~\ref{sec:grounding_details}.

\textbf{Learning To Propose}
Every proposal strategy has several hyperparameters, e.g.\ the temperature used for instruction generation and whether to ground the proposer with a data summary, program control flow, etc. Optimal configurations of these hyperparameters may depend in practice on the task, program, and proposer LM. For example, the dataset summary may be essential to a logical reasoning task but may distract a small proposer LM for highly familiar tasks like factoid question answering.  Motivated by this, in learning to propose, we parameterize proposal hyperparameters, and learn a Bayesian model over several trials to find what proposal strategy works for a given task, program, and LM setup.

\subsection{Credit Assignment}
Proposed assignments may be combined in many configurations. To search this space, we must identify the contribution of specific choices to LM program performance. We propose and explore three solutions for this credit-assignment problem: \textbf{greedy}, \textbf{surrogate}, and \textbf{history-based}. %

\textbf{Greedy}
As one technique, we consider proposing and evaluating single-stage changes to the LM program separately. This method limits the mis-attribution of errors to incorrect stages, but is inefficient as (i) changes must be applied one at a time, and (ii) changes to some stages may not change program-level performance until other stages are improved first.  Our preliminary experimentation with greedy credit assignment demonstrated that it was no more effective than other approaches but it imposed considerably worse time complexity.

\textbf{Surrogate}
To achieve more efficient credit assignment, we propose the use of Bayesian learning, which is known for it's ability to efficiently optimize functions with multiple latent variables. In this setup, a surrogate model learns to predict the quality of different parameter combinations from previous evaluations, allowing us to focus future exploration on the promising regions of the search space. In practice, we use Optuna's implementation of the Tree Structured Parzen Optimizer to build a Bayesian model over the quality of parameter combinations for the LM program \citep{akiba2019optuna, bergstra2011algorithms}. This multivariate variation of TPE models joint contributions between parameter choices, allowing us to jointly optimize our program's variables \cite{pmlr-v80-falkner18a}. In short, surrogate-based optimization allows us to optimize efficiently over a discrete set of existing parameter proposals. However, a shortcoming of this is that it only allows for optimization over a fixed set of proposals. Learnings from past evaluations cannot be used to improve the proposals themselves.

\textbf{History-Based} Here we make the assumption that given a history of past evaluations, a sufficiently strong LM could perform credit assignment, removing the need for an explicit surrogate model. This would allow us to perform credit assignment \textit{and} propose improved instructions simultaneously, as the proposer LM could in theory do both at once. Following the strategy outlined in OPRO \citep{yang2023large}, we rely on the proposer LM to learn assigned credit from a history of evaluated instructions and their scores by including these in the context. The proposer LM then outputs a new instruction based on this information. In Section~\ref{sec:optimizers}, we cover a few ways in which the LM can be instructed to conduct this credit assignment process more or less explicitly.

\section{Optimizers}
\label{sec:optimizers}

We now motivate specific instantiations of optimizer algorithms composed of the algorithmic strategies outlined in Section~\ref{sec:design}. %
We describe the algorithms that are the focus of our experimental investigation, and leave 
full descriptions for the remaining optimizers to Appendix~\ref{sec:algo_appendix}.

\subsection{Bootstrap Random Search}

\begin{figure}[H]
    \centering
    \includegraphics[width=0.96\linewidth]{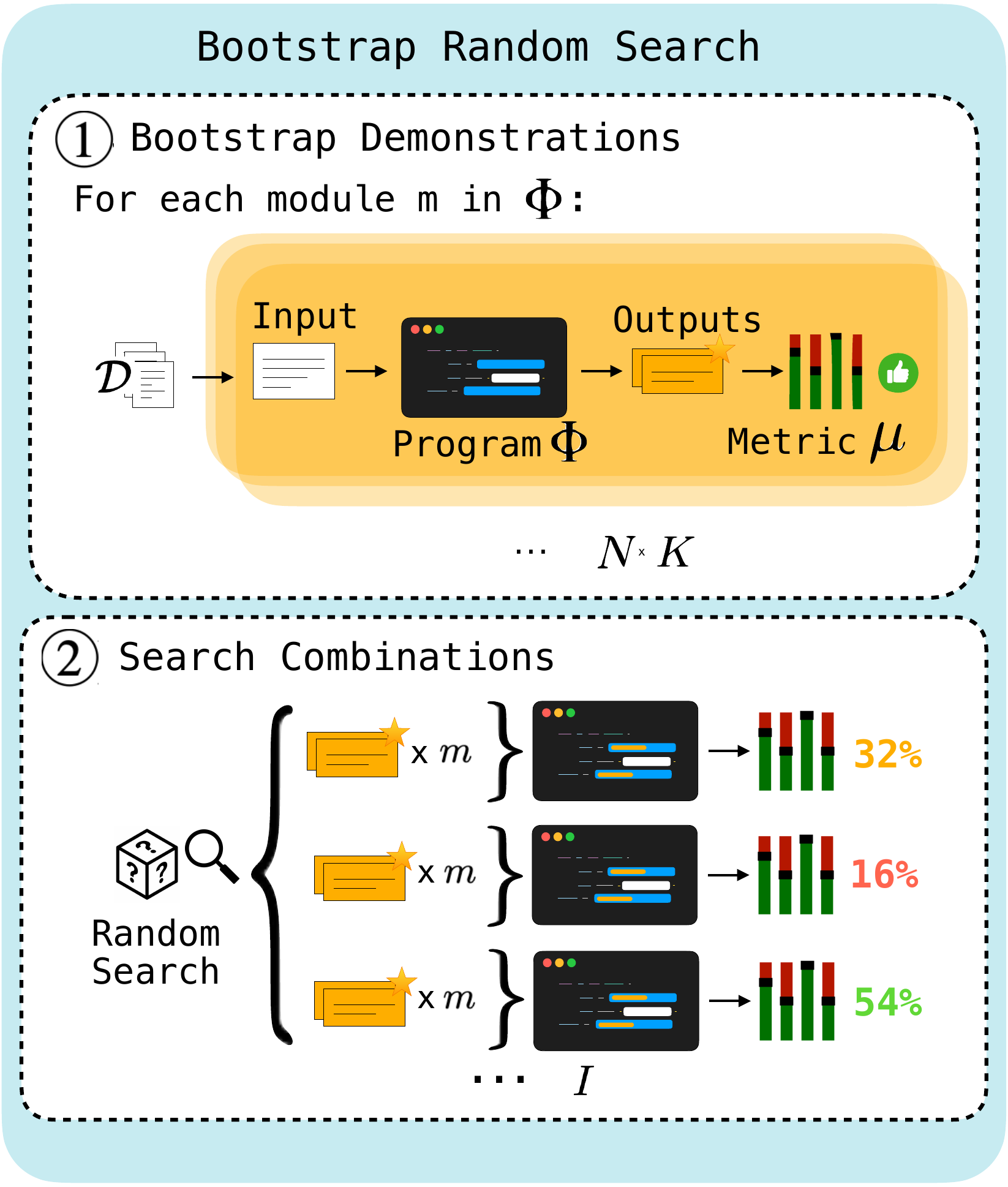}
    \caption{Bootstrap Random Search. In Step 1, demonstrations are bootstrapped by running training inputs through the program $\Phi$ and keeping traces that produce sufficiently high scoring outputs, as judged by metric $\mu$. In Step 2, these bootstrapped demonstration sets are searched over using random search, and the most performant set is returned.}
    \label{fig:bs_figure}
\end{figure}

\citet{khattab2024dspy} achieve strong results by generating and selecting task demonstrations for each module using random search. This approach fits into our general framework and serves as a strong baseline in our experiments. This optimization procedure (highlighted in Figure \ref{fig:bs_figure}) works as follows:

The hyperparameters include: $K$, the number of demonstrations to use for each module in $\Phi$, and $N$, the number of total sets to bootstrap and evaluate.
To \textbf{Initialize}, $N$ sets of few-shot examples are bootstrapped using the Bootstrapping Demonstrations procedure described in \ref{sec:prompt_proposal}. An input-output pair $(x, x') \in \mathcal{D}$ is randomly sampled from $\mathcal{D}$, where $x$ is an input to the program and $x'$ contains metadata (e.g.\ empty or final answers). Then, $\Phi(x)$ is run, which generates a full trace $\tau$ of the steps that $\Phi$ used for example $x$. If the output scores highly, i.e.\ $\mu(\Phi(x), x') \geq \lambda$ for some threshold $\lambda$ and given metadata or final label $x'$, we assume the trace to be correct, and use the inputs / outputs for each module in the trace as candidate few-shot examples. This process is repeated until $N$ sets of $K$ few-shot examples for each module have been bootstrapped.
To \textbf{Propose}, we sample a set of few-shot examples and use them to parameterize each module in $\Phi$.
To \textbf{Update}, the parameterized $\Phi$ is added to a global store along with its evaluation score on the training set (or a dedicated validation split thereof).
To \textbf{ExtractOptimizedSets}, the top-scoring assignment is returned.

\subsection{Module-Level OPRO}

\begin{figure}[H]
    \centering
    \includegraphics[width=0.95\linewidth]{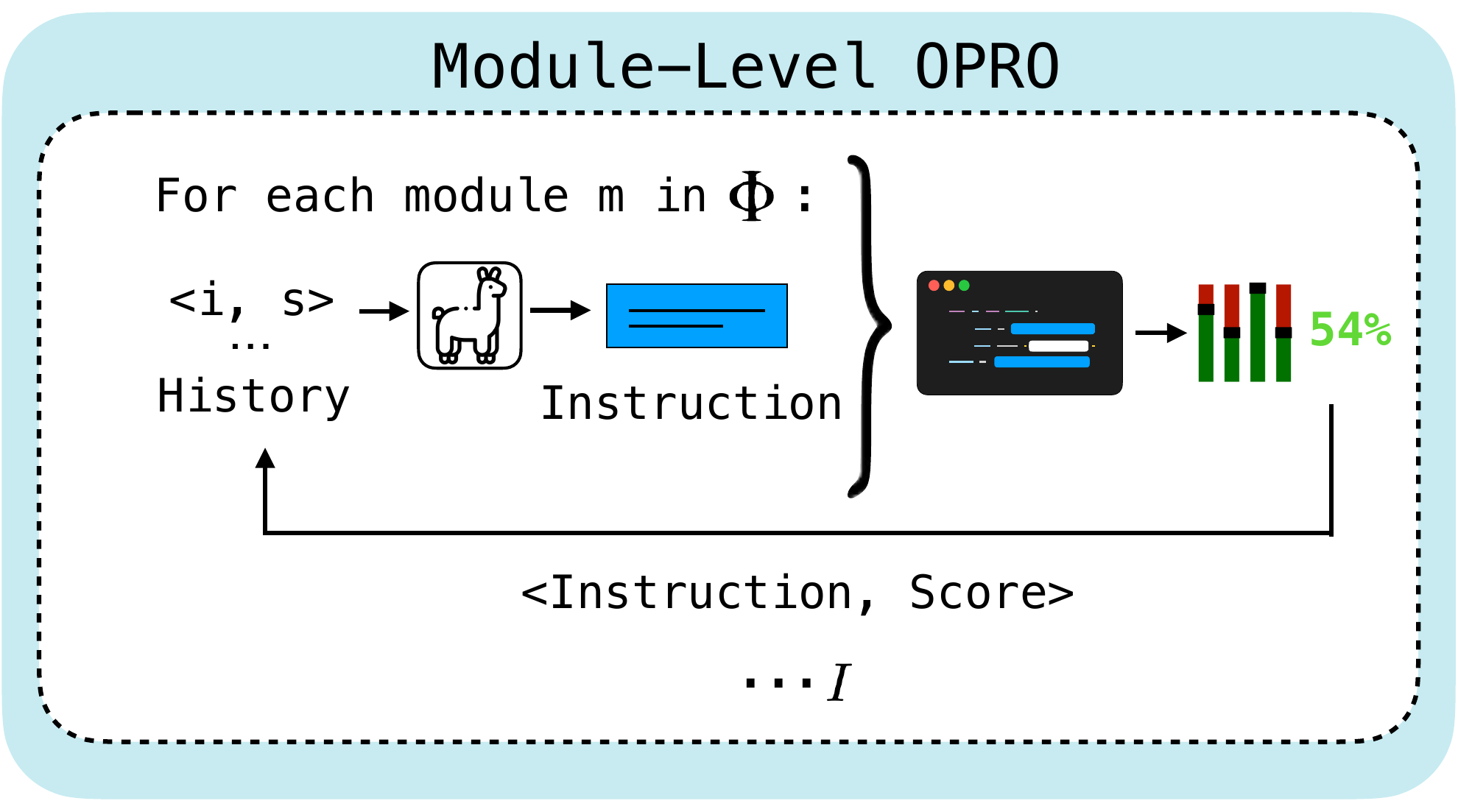}
        \caption{The Module-Level OPRO optimizer. A history of module-level instructions and program score pairs are given as input to the proposer LM to generate a new instruction for each module. These are then evaluated in the program, and the resulting score is added back with each module's instruction to the module's history. The process repeats for $I$ iterations.}
    \label{fig:opro_figure}
\end{figure}

We seek to extend the OPRO algorithm \citep{yang2023large} to an arbitrary LM program $\Phi$, e.g.\ with $m \geq 1$ stages embedded in a larger program. In OPRO, the proposer LM is provided with a history of proposed instructions and their scores, allowing it to learn to propose better instructions over time. To extend OPRO, we first consider an approach we refer to as Module-Level OPRO, in which we assume that the program score is a good enough proxy for an individual instruction's quality. In other words, even though the program is parameterized with $m$ instructions, we will assume that the score is reflective enough of each. Module-level OPRO (Figure \ref{fig:opro_figure}) works as follows: 

To \textbf{Initialize}, a seed instruction is used to parameterize each module in $\Phi$, which is evaluated. To \textbf{Propose}, the resulting score and the $i_{th}$ module's instruction are inputted into the proposer LM to create a new instruction for module $i$. This is done for all $m$ modules. The set of $m$ generated instructions are then used to parameterize $\Phi$, and the parameterized program is evaluated. To \textbf{Update}, each instruction and the score is added to the history for each module. The OPRO sub-routine is run again with the updated histories to create a new set of instructions. We note that optimizers for each stage are provided with histories of module-level trajectories and proxy scores only. This process continues until the maximum number of iterations is reached. To \textbf{Extract\-OptimizedSets}, the parameterization of $\Phi$ that scored highest is returned.

\subsection{MIPRO}
\label{mipro_section}

\begin{figure}[H]
    \centering
    \includegraphics[width=0.97\linewidth]{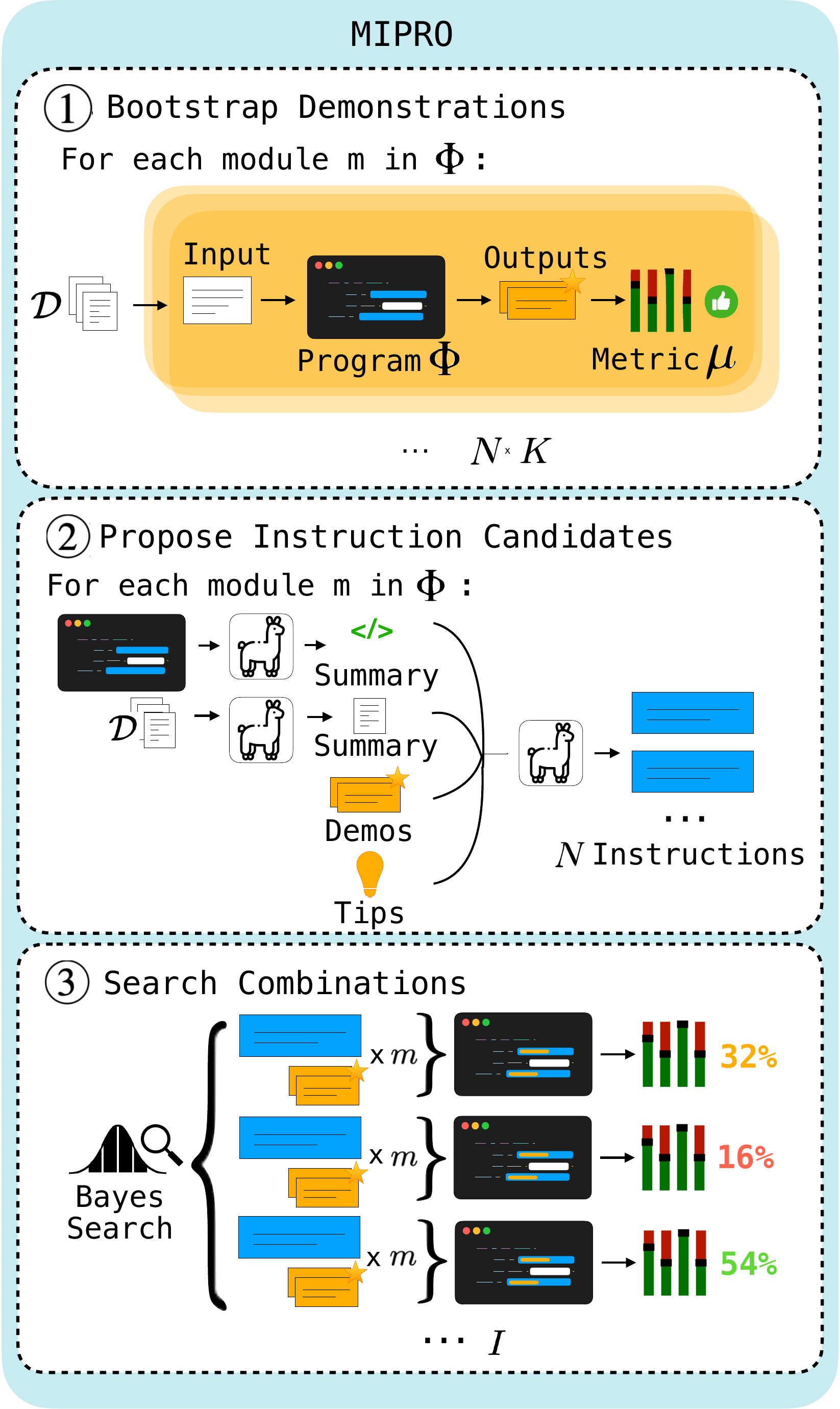}
        \caption{The MIPRO optimizer. In Step 1, demonstrations are bootstrapped using the same process from Step 1 of Bootstrap Random Search. In Step 2, instructions are proposed using the grounding strategy described in \ref{sec:prompt_proposal}. In Step 3, Bayesian optimization is used to find the best performing combination of instruction and demonstration candidates.}
    \label{fig:MIPRO_figure}
\end{figure}
To relax OPRO's strong assumptions, we propose the use of a Bayesian surrogate model to explicitly learn the sensitivity of task-level scores to module-level parameters such as instructions and demonstrations throughout optimization. We refer to this approach as MIPRO (\underline{M}ulti-prompt \underline{I}nstruction \underline{PR}oposal \underline{O}ptimizer). By separating the task of credit assignment from prompt proposal, we allow for the proposal LM to focus on the task of proposal only. Credit assignment and final selection is then done post-hoc using our surrogate model.

Bayesian optimization is known for its robustness to noise, as it effectively incorporates uncertainty into the optimization process \cite{snoek2012practical}. We therefore propose evaluating over mini-batches of our training data, rather than the full set with each iteration. This allows us to explore and exploit parameter configurations more efficiently. The MIPRO algorithm (Figure \ref{fig:MIPRO_figure}) is as follows:

To \textbf{Initialize}, MIPRO bootstraps a set of $N$ few-shot example sets and instructions per module using the Bootstrap Demonstration and Grounding strategies respectively, found in \ref{sec:prompt_proposal}. Latent categorical variables representing the choice of few-shot examples and instructions for each module are initialized with a uniform prior. To \textbf{Propose}, we use the sampling rule from the Tree-structured Parzen Estimator \citep{bergstra2011algorithms} to select the instructions and few-shot examples used to parameterize $\Phi$. To \textbf{Update}, this parameterized $\Phi$ is scored on a randomly selected mini-batch of B samples, and the scores are used to update the Estimator's priors over parameter quality. To \textbf{ExtractOptimizedSets}, every $S$ steps, the candidate parameterizations of $\Phi$ with the highest mean score over trials is evaluated on the full train-set. At the end, the highest scoring fully evaluated parameterization is returned as the optimal assignment.

\subsection{Other MIPRO variants}

\hspace*{\parindent}
\textbf{0-Shot MIPRO} is a straightforward extension of MIPRO that simply optimizes over instructions only, rather than both jointly. This could be desirable for cost or context-window constraints.

\textbf{Bayesian Bootstrap} is the restricted version of MIPRO to optimizing over bootstrapped demonstrations. This may be ideal when few-shot examples are essential to the task or when we have already identified a good instruction and want to use our budget to optimize demos alone.

\textbf{MIPRO++} applies a Bayesian surrogate model to optimize \textit{proposal} hyperparameters, rather than the choice of LM program parameters themselves. This follows directly from the Learning to Propose strategy discussed in Section~\ref{sec:prompt_proposal}. In the full form of this approach, a surrogate  model is used to learn optimized parameters for proposing instructions, as well as for bootstrapping demonstrations. However, in the context of this work, we focus on evaluating this approach for optimizing our instruction proposal strategy specifically (described next). 

\textbf{0-Shot MIPRO++} tunes instructions by meta-optimizing how or whether our Grounded instruction proposal strategy uses the dataset summary (\texttt{boolean}), uses the program summary (\texttt{boolean}), adjusts the proposer LM temperature (\texttt{float}), provides the proposer LM a plain-text tip on prompt engineering  (\texttt{categorical}; see Appendix~\ref{sec:grounding_details}), and selects a specific set of bootstrapped demos to show to the proposer LM (\texttt{categorical}). A Bayesian model with the same mini-batching strategy employed in MIPRO is then used to optimize over these hyperparameters. A program with the newly proposed instruction is evaluated on each trial, using the mini-batching approach described earlier. The best fully-evaluated program is returned.

\begin{table*}[tp]
\begin{center}
\resizebox{\textwidth}{!}{%
\begin{tabular}{lllccl}
\toprule
\textbf{Benchmark} & \textbf{Task Type} & \textbf{Program} & \textbf{Modules} & \textbf{LM Calls} & \textbf{Metric} \\ \midrule
HotPotQA & Multi-Hop QA & Multi-Hop Retrieval & 2 & 3 & Exact Match \\
HotPotQA Conditional & Multi-Hop QA & Multi-Hop Retrieval & 2 & 3 & Custom \\
Iris & Classification & Chain of Thought & 1 & 1 & Accuracy \\
Iris-Typo & Classification & Chain of Thought & 1 & 1 & Accuracy \\
Heart Disease & Classification & Answer Ensemble & 2 & 4 & Accuracy \\
ScoNe & Natural Language Inference & Chain of Thought & 1 & 1 & Exact Match \\
HoVer & Multi-Hop Claim Verify & Multi-Hop Retrieval & 4 & 4 & Recall@21 \\
\bottomrule
\end{tabular}
}
\caption{DSPy Optimizer Benchmark and associated programs. We benchmark our optimizers on seven diverse programs. Additional details are in Appendices~\ref{sec:task_desc} and~\ref{sec:lm-program-appendix}.}
\label{tab:optimizer-benchmark}
\end{center}
\end{table*}

\subsection{Other OPRO variants}

\hspace*{\parindent}
\textbf{Program-Level OPRO} provides the proposer LM with a history of \textit{full, multi-stage} trajectories and relies on it to assign credit for task-level scores to the program stages. While Module-level OPRO embeds the assumptions that there is no inter-assignment dependency and that credit assignment across modules is equal, Program-level OPRO assumes that an LM will successfully complete credit-assignment when provided with long trajectory histories. These are all very strong assumptions. In particular, information contained in histories is likely to be lost as history length grows \cite{liang_litm}. In our experiments, we opt for using Module-level OPRO because Program-Level OPRO is more complex and did not appear to provide additional performance gains.

\textbf{CA-OPRO} ``Coordinate-Ascent'' OPRO (CA-OPRO) employs a greedy credit assignment approach to extend OPRO to multi-stage settings. It iterates through each module $m$ in the program, proposes a new set of $N$ instructions for $m$ using Module-Level OPRO's proposal function, and evaluates each proposal by keeping all other parameters in the program fixed. It then updates module $m$ with the best evaluated instruction so far, and repeats with the next module. This entire process is then repeated $D$ times. From initial experiments, we found CA-OPRO's performance did not justify its inefficiency, so we focus our experiments on evaluating other methods more thoroughly.

\section{Experimental Setup}
\label{sec:experiments}

\subsection{Benchmark}\label{sec:benchmark}
We develop seven tasks (i.e.\ seven groups of dataset, metric, and LM program) to evaluate LM program optimizers. Table~\ref{tab:optimizer-benchmark} presents our tasks, whose full descriptions, splits, and DSPy program pseudocode are presented in Appendices~\ref{sec:task_desc}, \ref{sec:data_splits} and~\ref{sec:lm-program-appendix}, respectively. We use 500 examples for training, 500 for our development, and a test set of 2k examples (or the full test set if smaller).

As Table~\ref{tab:optimizer-benchmark} shows, we include four multi-stage and two single-stage programs. \textbf{HotPotQA}~\cite{yang-etal-2018-hotpotqa}, in the ``fullwiki'' setting, requires systems to answer factoid questions by retrieving two relevant articles from Wikipedia. We build a program with a module for generating search queries (invoked twice) and another for generating the final answer. This is a canonical test of LM program optimizers, based on~\citealt{khattab2024dspy}. 

We hypothesize that optimizing \textit{free-form instructions} can have the most impact on tasks with \textit{subtle} rules that \textit{cannot be properly inferred through a few examples}. We devise \textbf{HotPotQA Conditional} to test this: we change the answer format accepted from the program depending on whether the answer is a person, a date, a place, etc.

We also include two classical classification tasks: \textbf{Iris} (\citealt{Fisher1936THEUO}; flower classification given six real-numbered features) and \textbf{Heart Disease} (\citealt{Detrano1989InternationalAO}; binary classification given 13 categorical and continuous features). We test Iris in two settings, one with a misspelling in the prompt that may confuse the LM and one corrected.\footnote{The misspelling was initially accidental (asking to classify a flower as ``versicolour'' rather than ``versicolor''). However, it serves as a realistic test for optimization from a misspecified prompt, so we report on both settings. Future work should permit LM programming abstractions to detect such errors.} Iris can be nearly solved with a simple set of rules (not provided to the LM), so we seek to test if LM program optimizers can automatically teach LMs a Chain-of-Thought behavior to also perform well on such tasks. In contrast, it may be harder to find a small number of crucial patterns in Heart Disease, and we thus test a  program that generates three clinical opinions using Chain-of-Thought LM calls and then generates a final judgment accordingly.

To assess whether optimizers can express data-specific nuances that are not evident from the program itself, we use \textbf{ScoNe}~\citep{she-etal-2023-scone}, an entailment task in which LMs must reason about logical puzzles with nested negation. %

Lastly, we evaluate three-hop retrieval over unchecked claims using \textbf{HoVer}~\citep{jiang-etal-2020-hover}. Our LM program alternates three times between generating queries, using them for retrieval from Wikipedia, and using the results to inform future queries. We use HoVer's gold labels for the documents to be retrieved for each input claim to report Retrieval@21 with all top-10 retrieved documents across three hops.

\subsection{Methods \& Models}

We evaluate the optimizers discussed above for (i) instruction-only optimization, (ii) few-shot optimization, and (iii) joint instruction \& few-shot optimization. For instruction-only optimization, we compare Module-Level OPRO, 0-Shot MIPRO, and 0-Shot MIPRO++. For optimizing few-shot demonstrations only, we compare Bootstrap Random Search (RS) with Bayesian Bootstrap. For optimizing both instructions and few-shot demonstrations together, we use MIPRO. We use an un-optimized LM Program as a baseline. In order to evaluate the utility of Grounding, we compare Module-Level OPRO with a version without Grounding, which we refer to as Module-Level OPRO $-$G. In these experiments, we use only the components described in the original OPRO paper in our proposal prompt: few-shot examples, and a history of previously proposed instructions and their scores. Optimizers are run for a budget of 20--50 trials with full evaluation on the trainset, depending on the task. Note that this translates to a larger number of actual optimization trials for optimizers using minibatching. More information on the exact budgets used for each task, as well as experiment hyperparameters, are detailed in Appendix~\ref{experiment_hyperparameters}. We conduct 5 runs for each method on each task. We use Wilcoxon signed-rank  tests between the averages of all runs for each example in the test set to help assess the statistical significance of performance differences between two methods.

In the majority of experiments, we use GPT-3.5 as our proposer LM (the model that crafts instructions) with a default temperature of 0.7, and Llama-3-8B as our task model (the LM used inside the LM programs). We note that the instruction proposal temperature is updated as a hyperparameter in our Learning to Learn experiments. For bootstrapping few-shot demonstrations, we use Llama-3-8B as a default teacher-model, but switch to GPT-4o for more challenging tasks (ScoNe and HoVer).

\section{Results \& Discussion}
\label{sec:results_and_discussion}

\begin{table*}[h]
\centering
\setlength{\tabcolsep}{1.5pt}
\resizebox{\textwidth}{!}{
\begin{tabular}{>{\centering\arraybackslash}m{4cm} | *{3}{>{\centering\arraybackslash}m{1.0cm}} | *{3}{>{\centering\arraybackslash}m{1.0cm}} | *{3}{>{\centering\arraybackslash}m{1.0cm}} | *{3}{>{\centering\arraybackslash}m{1.0cm}} | *{2}{>{\centering\arraybackslash}m{1.0cm}} | *{2}{>{\centering\arraybackslash}m{1.0cm}} | *{2}{>{\centering\arraybackslash}m{1.0cm}}}
\toprule
Optimizer & \multicolumn{3}{c|}{ScoNe} & \multicolumn{3}{c|}{HotPotQA} & \multicolumn{3}{c|}{HoVer} & \multicolumn{3}{c|}{HotPotQA Cond.} & \multicolumn{2}{c|}{Iris} & \multicolumn{2}{c|}{Iris-Typo} & \multicolumn{2}{c}{Heart Disease} \\
& Train & Dev & Test & Train & Dev & Test & Train & Dev & Test & Train & Dev & Test & Train & Test & Train & Test & Train & Test \\
\midrule
\multicolumn{19}{l}{\textit{Instructions only (0-shot)}} \\
\midrule
N/A & 57.0 & 56.2 & 69.1 & 35.4 & 31.8 & 36.1 & 30.2 & 30.8 & 25.3 & 13.8 & 10.5 & 6 & 46.4 & 40.9 & 34.7 & 32 & 23.3 & 26.8 \\
Module-Level OPRO $-$G & 70.0 & 67.4 & 76.1 & 36.0 & 31.7 & 36.0 & 30.0 & 30.0 & 25.7 & -- & -- & -- & -- & -- & -- & -- & -- & -- \\
Module-Level OPRO & 69.1 & 67.6 & 73.5 & 41.9 & 36.2 & 39.0 & 37.1 & 38.6 & 32.5 & -- & -- & -- & -- & -- & -- & -- & -- & -- \\
0-Shot MIPRO & 66.3 & 65.2 & 71.5 & 40.2 & 34.2 & 36.8 & 37.7 & 38.4 & 33.1 & 22.6 & 20.3 & 14.6 & 40.8 & 36.4 & 56.8 & 56.7 & 26.8 & 25.8 \\
0-Shot MIPRO++ & 69.0 & 66.9 & 75.7 & 41.5 & 36.2 & 39.3 & 37.1 & 37.3 & 32.6 & -- & -- & -- & -- & -- & -- & -- & -- & -- \\
\midrule
\multicolumn{19}{l}{\textit{Demonstrations only (Few-shot)}} \\
\midrule
Bootstrap RS & 74.9 & 69.6 & 75.4 & 48.6 & 44.0 & \textbf{45.8} & 42.0 & 42.0 & 37.2 & 16.4 & 15.0 & 10.4 & 95.2 & \textbf{94.1} & 58.9 & 58.7 & 78.4 & \textbf{79.2} \\
Bayesian Bootstrap & 75.4 & 67.4  & 77.4  & 49.2 & 44.8 & \textbf{46.2} & 44.6 & 44.7 & 37.6 & -- & -- & -- & -- & -- & -- & -- & -- & -- \\
\midrule
\multicolumn{19}{l}{\textit{Both (Few-shot)}} \\
\midrule
MIPRO & 74.6 & 69.8 & \textbf{79.4} & 49.0 & 43.9 & \textbf{46.4} & 44.7 & 46.7 & \textbf{39.0} & 28.4 & 28.1 & \textbf{23.3} & 98.4 & 88.6 & 69.1 & \textbf{68.7} & 75.2 & 74.2 \\
\bottomrule
\end{tabular}}
\caption{Results averaged across 5 runs, divided into optimizing instructions only (i.e., ``0-shot'' prompts), demonstrations only, and both. The best performing values in each column are highlighted in bold. These bold values represent the highest average scores compared to the second-highest, with significance supported by Wilcoxon signed-rank tests (p < .05) between the corresponding run averages. If significance is not confirmed, multiple results are bolded to denote their comparable performance.}
\vspace{-3mm}
\label{table:performance_comparison}
\end{table*}

Table~\ref{table:performance_comparison} summarizes our main results, from which we derive five overarching lessons.

\textbf{Lesson 1: Optimizing bootstrapped demonstrations as few-shot examples is key to achieving the best performance.} For the majority of tasks, we find that optimizing boostrapped demonstrations alone yields significantly better performance than optimizing instructions alone. We confirm this with a Wilcoxon signed-rank statistical test, which shows that even simple Bootstrap Random Search beats the best instruction-only optimizer for a given task in all but one case. The exception to this is HotPotQA Conditional, which supports our hypothesis and findings discussed in Lesson 3. We finally note that creating the \textit{right} set of bootstrapped demonstrations is important. Our optimization runs  indicate that there is high variation in the performance resulting from different few-shot sets (see Appendix~\ref{sec:optimization_plots}). We thus infer that strong bootstrapped examples provide information pertaining to successful reasoning behavior more than just, say, teaching task format.

\textbf{Lesson 2: Optimizing both instructions and few-shot examples  with MIPRO generally yields the best overall performance.} 
We support this with a statistical test comparing MIPRO with the second highest averaging optimizer for each task. The exceptions to this are HotPotQA, Heart Disease, and Iris without a typo. We hypothesize that instructions are less valuable for tasks like HotPotQA, whose final module is a relatively straightforward Q\&A task that is likely in-distribution for many models. For Heart Disease, we hypothesize that this is due to initializing our optimizers with a simple seed instruction that does not convey any classification criteria, which current instruction optimizers have a limited ability to infer. 

\textbf{Lesson 3: Instruction optimization is most important for tasks with conditional rules that are (i) not immediately obvious to the LM and (ii) not expressible via a limited number of few-shot examples.} This hypothesis is supported by our Iris-Typo experiment---and primarily by our HotPotQA Conditional experiments, where we optimize over a seed instruction stating the rules of the task. For this task, we find that even 0-shot instruction optimization outperforms demonstration-only optimization. In these cases, especially if the task is complex, it's important that we optimize over a seed prompt, as our optimizers are not yet able to infer all task rules. (This lesson is also reflected in the Heart Disease results, as discussed above.)  In the Iris-Typo setting, our instruction optimizer even helps correct mistakes in the seed prompt. 

\textbf{Lesson 4: Grounding is helpful for instruction proposal overall, but the best proposal strategy varies by task.} 
In our Module-Level OPRO Grounding ablations, we find that Grounding is essential for performance improvements for HotPotQA and HoVer, but seems to hurt performance for ScoNe. This motivates approaches like MIPRO++, which are able to learn custom proposal strategies for a given task. Indeed, we see that 0-Shot MIPRO++ is able to learn a proposal strategy that recovers this performance for ScoNe. One additional benefit of 0-Shot MIPRO++ is that the learned importance scores from the Bayesian model used to optimize proposal hyperparameters can provide us insight into the utility of each proposal component. Studying these importance scores (found in Appendix~\ref{sec:param_importance}) reveals, across tasks, the highest importance scores go to the choice of bootstrapped demonstrations in the meta-prompt and the tip. We observe that the importance of many parameters varies between tasks: for example, the dataset summary has a high learned importance score for ScoNe, whereas it is one of the least important parameters for HotPotQA and HoVer.

\textbf{Lesson 5: There is more to learn about LM program optimizers.} When comparing the performance of Module-Level OPRO, 0-Shot MIPRO, and 0-Shot MIPRO++, we find that results are mixed. Bayesian Bootstrap outperforms Bootstrap Random Search for ScoNe, but this finding is not statistically significant for HotPotQA and HoVer. 0-shot MIPRO++ outperforms 0-shot MIPRO for ScoNe and HotPotQA, but is about equivalent in the case of HoVer. Future work may find more differentiated results when studying these optimizers at different optimization budgets: for example, it's possible that 0-shot MIPRO would perform best in very low budget settings, given that it's use of minibatching allows it to explore many more parameter options with the same budget. Conversely, 0-shot MIPRO++ may shine in scenarios where budget is not an issue, and spending many trials to learn optimal proposal dynamics could lead to differentiated results. We save this exploration for future work.

\section{Related Work}
Recent work has explored optimizing string prompts, 
including gradient-guided search \citep{shin-etal-2020-autoprompt, wen2023hard}, reranking brute force search \citep{gao-etal-2021-making}, evolutionary algorithms \citep{fernando2023promptbreeder}, prompting other LMs \citep{yang2023large, zhou2023large, pryzant2023automatic}, and reinforcement learning (RL) \citep{deng-etal-2022-rlprompt, zhang2022tempera, hao2022optimizing}.  Prior work on RL for prompts focuses on word level edits of only a few words \citep{deng-etal-2022-rlprompt}, phrase level edits \citep{zhang2022tempera}, or text-to-image generation \citep{hao2022optimizing}. 
\citet{khattab2024dspy}  present DSPy, a programming model for expressing LM programs and optimizing their prompts 
and weights. Unlike our work, the authors only explore optimizers based on bootstrapping strong demonstrations. \citet{sordoni2023joint} explore joint prompt optimization for stacked LLM calls, modeling this as variational inference and exploring this for two simple layers. Their approach inherently relies on having access to log probabilities for explicitly passed tokens to LMs, an increasingly restrictive assumption in practice (e.g.\ with commodified LM APIs that allow only text-in-text-out processing). In contrast, our optimizers work on an arbitrary number of modules for any LM program out-of-the-box.

\section{Conclusion} %
We formalize the problem of optimizing prompts in LM programs (\S \ref{sec:problem-statement}).  We identify two key challenges of LM program prompt optimization: (1) proposal of a small set of high-quality prompts and (2) credit assignment during optimization.  We address these challenges using three proposal generation and three credit assignment strategies (\S \ref{sec:design}) and explore a representative subset of optimizer algorithms (\S \ref{sec:optimizers}) using a new benchmark of diverse tasks (\S \ref{sec:benchmark}). Our findings show that optimizing few-shot demonstrations is very powerful, but instruction optimization can be essential for complex task specifications with  multiple conditional rules. Finally, we find that jointly optimizing both instructions and demonstrations using the MIPRO optimizer is the most effective approach in five out of seven settings.

\section*{Limitations}
This work studies a set of optimizers under a fixed budget, but does not examine how optimization dynamics might differ across extremely low or high budget scenarios. As discussed in Section~\ref{sec:results_and_discussion}, doing so may reveal new insights about the trade-offs between different optimizers, such as those that learn to improve proposals overtime, versus those that optimize over existing proposals very efficiently.
In our experiments, we also use a fixed proposer LM and task LM. Future research should assess whether the proposed methods demonstrate consistent performance when employing different models. 
Furthermore, a limitation of the optimizers introduced in this work is their restricted ability to infer the rules governing complex tasks without a handwritten seed prompt. While some information can be gleaned from grounding—such as dataset details or examples of the task—this may be insufficient for extrapolating a comprehensive set of rules. We encourage subsequent studies to investigate how optimizers could learn such task dynamics without relying on handwritten inputs.
Finally, while we have made efforts to establish a benchmark that covers a diverse range of tasks and programs, there remains more to learn regarding the performance of optimizer methods on increasingly complex tasks and programs. Improving this benchmark presents a promising avenue for future research.

\section*{Acknowledgements}

Krista Opsahl-Ong is supported by the NSF CS-Grad4US Graduate Fellowship.  This work was partially supported by IBM as a founding member of the Stanford Institute for Human-Centered Artificial Intelligence (HAI), and by the HAI Hoffman–Yee Grant “Dendritic Computation for Knowledge Systems”. Additional support was provided by the Wallenberg AI, Autonomous Systems and Software Program (WASP), funded by the Knut and Alice Wallenberg Foundation, as well as by Digital Futures at KTH. Finally, this research was also supported in part by affiliate members and other supporters of the Stanford DAWN project -- Facebook, Google, and VMware.

\bibliography{custom}

\appendix

\clearpage
\section{Detailed Task Descriptions}
\label{sec:test_train_split}
\label{sec:task_desc}

\textbf{HotPotQA} \cite{yang-etal-2018-hotpotqa} is a dataset of  questions that require reasoning over multiple Wikipedia articles to answer. We adopt the ``fullwiki'' setting in which the system must retrieve the right articles from all 5M Wikipedia page abstracts. Our LM program, derived from \citealt{khattab2024dspy}, involves three stages, given a question: (1) generating a search query that a retrieval model uses to find Wikipedia passages, (2) reading those passages to generate a second ``hop'' query for the retrieval model, and finally (3) reading all of the retrieved passages to answer the question. %

\textbf{HotPotQA Conditional} We hypothesize that optimizing instructions can have the most impact on tasks with rules that (1) are not immediately obvious to the task LM and (2) cannot be fully defined through a few examples. To test this, we devise a task that applies additional conditional rules to HotPotQA: depending on the category of the answer, the LM must respond in a specific format:
\begin{itemize}\setlength{\itemsep}{0pt}
\item When the answer is a person, the response must be in lowercase. When it's a place, the response should contain no punctuation. 
\item When the it's a date, the response must end with ``Peace!'', but in no other circumstances. 
\item If answer falls into another category, the response must be in all caps.
\end{itemize}
We use GPT-4 to annotate the answer types as ``person'', ``place'', or ``date'' with author supervision for applying special conditional rules based on answer type. We use the same multi-hop LM program described above to solve this task, and initialize the program with an instruction describing the rules of the task. This program is then evaluated using exact match along with regex parsing to determine if all other conditions are being followed.

\textbf{Iris} is a classic classification dataset over floating-point features~\citep{Fisher1936THEUO}. It involves classifying a flower as one of three iris species given the flower's sepal and petal width and length.  Our LM program for this task is a simple Chain-of-Thought program, initialized the instruction: ``Given the petal and sepal dimensions in cm, predict the iris species.'' Crucially, we know that Iris can be nearly solved with a simple set of rules (not provided to the LM) and seek to test if LM program optimizers can automatically teach LMs to also perform well on such tasks.

\textbf{Heart Disease} \cite{Detrano1989InternationalAO} is a classification task where it may be harder to find a small number of crucial patterns. Given a set of 13 features including a patient's age, biological sex, and cholesterol, we must predict whether the patient has heart disease. Our LM program generates three separate clinical opinions using Chain-of-Thought LM calls and then generates a final judgment based on the opinions given.

\textbf{ScoNe} \citep{she-etal-2023-scone} evaluates logical reasoning with negation. We define a simple Chain-of-Thought program to reason over ScoNe entailment questions and produce a binary answer. %
ScoNe allows us to represent rich single-stage tasks in our evaluation. We hypothesize that ScoNe's focus on NLI and logical deduction with nested negations evaluates whether optimizers can express task nuances via instructions or demonstrations.

\textbf{HoVer} \citep{jiang-etal-2020-hover} contains claims that require many-hop search steps over Wikipedia to be fact-checked. We consider the sub-task of retrieving all required evidence. Our LM program performs a search on the claim, summarizes the results (LM call 1), adds that to context and proposes the next search query (LM call 2), performs a search, summarizes the results (LM call 3), proposes a search query based on both summaries (LM call 4), and finally searches once more. HoVer has gold labels for the documents to be retrieved. We measure Retrieval@21 with all top-10 retrieved documents across three hops. We restrict HoVer to examples with 3 supporting facts in order to study optimization in a more challenging setting. We note that different supporting facts can originate from the same document, so finding all required documents can sometimes be done in <3 hops. Approximately 67\% (83\%) of the training (test) sets are 3-hop queries, and the rest are 2-hop queries. In follow-up studies, we recommend filtering by examples that require 3-hops to study retrieval in a more uniform setting.

\section{Experiment Setup Details}
\label{experiment_hyperparameters}

\subsection{Data Splits}\label{sec:data_splits}

Our data splits are described in Table~\ref{tab:test_train_split}.  We generally use 500 examples for training, 500 for our development, and a test set of 2k examples (or the full test set for tasks with test sets <2k samples).

\begin{table}[tp]
\begin{center}
\small
\begin{tabular}{lccc}
\toprule
\textbf{Benchmark} & \textbf{Train} & \textbf{Dev} & \textbf{Test} \\ \midrule
HotPotQA & 500 & 500 & 2000 \\
HotPotQA Conditional & 500 & 200 & 200 \\
Iris & 75 & N/A & 75 \\
Heart Disease & 120 & N/A & 183 \\
ScoNe & 500 & 500 & 1200 \\
HoVer & 500 & 500 & 1520 \\
\bottomrule
\end{tabular}
\caption{Train, Dev, Test splits for each of our datasets. Training data was used for training our Optimizers. Dev data was used as a development set to internally iterate on methods. Test was used for final evaluation and reporting. The datset for HotPotQA Conditional is smaller because the labels were created by hand. We do not create a devset for Iris or Heart Disease due to the fact that these were (1) small datasets and (2) we did not use them for method iteration.}
\label{tab:test_train_split}
\end{center}
\end{table}

\subsection{Optimizer Budget}
We optimize HotPotQA and ScoNe with a budget of 50 full evaluation trials (which translates to about 300 minibatch trials).
Iris, Heart Disease, and HotPotQA Conditional are run with a budget of 30 full evaluation trials, and HoVer 20 full evaluation trials. We use less trials for HoVer experiments given that it is the most expensive program to run. We use less trials for Iris, Heart Disease, and HotPotQA Conditional given that these experiments were focused on understanding the value of instruction versus few-shot optimization rather than evaluating specific methods, which may need more trials to see differentiated results.
\subsection{Optimizer Hyperparameters}
Table~\ref{tab:opt_hyperparams} shows the number of instruction and / or few-shot demonstration candidates for each module ($N$) that 0-Shot MIPRO, Bayesian Bootstrap, and MIPRO were used to optimize over in our experiments. We note that these hyperparameters were not chosen with extensive sweeps, but were instead chosen built on intuition built over running past experiments. Our general rule of thumb was to set $N$ to be < $T/v$, where $T$ is the trial optimization budget, and $v$ is the total number of variables we are optimizing over.

\subsection{Language Model Hyperparameters}
We perform most of our experiments with the LLama 3 8B model \cite{llama3modelcard} which we serve using SGLang \cite{zheng2024sglang} on A100 GPUs.  We parallelize inference calls across 8 A100 GPUs for many of our experiments. However, only a single GPU capable of running Llama 3 8B or a cloud inference provider would be necessary to replicate all results.  The temperature of our Llama model is optimized as a part of the MIPRO++ experiments, but we otherwise use temperature 0.7.  We also always use top\_p=1.0 sampling for Llama.  We generate until the model reaches the max tokens for a given task or the stop tokens \texttt{\["\backslash n \backslash n", "\backslash n---", "assistant"\]}.  For ScoNe this is 200 tokens, for HoVer this is 600 tokens, and for all other tasks this is 150 tokens.  Importantly we run Llama 3 without a chat template as we find this behaves better given DSPy's autocomplete prompt style.  For our proposer model, we use GPT-3.5 with temperature 0.7 and top\_p=1.0 or GPT-4 with the same settings for ScoNe, HoVer, and Iris.

\begin{table}[tp]
\begin{center}
\small
\begin{tabular}{llc}
\toprule \textbf{Optimizer} &
\textbf{Benchmark} & \textbf{N} \\ \midrule
\multirow{6}{*}{0-Shot MIPRO} & HotPotQA & 60  \\
 & HotPotQA Conditional & 35 \\
 & Iris & 50 \\
 & Heart Disease & 30  \\
 & ScoNe & 70 \\
 & HoVer & 15 \\
 \midrule
\multirow{6}{*}{Bayesian Bootstrap} & HotPotQA & 60  \\
 & HotPotQA Conditional & N/A \\
 & Iris & N/A \\
 & Heart Disease & N/A  \\
 & ScoNe & 70 \\
 & HoVer & 15 \\
 \midrule
\multirow{6}{*}{MIPRO} & HotPotQA & 30  \\
 & HotPotQA Conditional & 30 \\
 & Iris & 30 \\
 & Heart Disease & 15  \\
 & ScoNe & 70 \\
 & HoVer & 10 \\
\bottomrule
\end{tabular}
\caption{Number of candidates per module (N) used for optimization. Note that this hyperparameter only applies to our 0-Shot MIPRO, MIPRO, and Bayesian Bootstraping optimizers, because for other optimizers the \# of candidates explored equals the number of trials.}
\label{tab:opt_hyperparams}
\end{center}
\end{table}

\section{Grounding Details}
\label{sec:grounding_details}

The following section includes details regarding the grounded prompting strategy used in our methods.

\subsection{Instruction Proposal Program}
Below is the signature for an LM program we use to generate instruction candidates. Note that a separate module is used to generate the dataset description and the program description.
\begin{lstlisting}[language=Python, caption={Instruction Proposal Program},captionpos=b]
class GenerateSingleModuleInstruction(dspy.Signature):
        (
            """Use the information below to learn about a task that we are trying to solve using calls to an LM, then generate a new instruction that will be used to prompt a Language Model to better solve the task."""
        )
        dataset_description = dspy.InputField(desc="A description of the dataset that we are using.",)
        program_code = dspy.InputField(desc="Language model program designed to solve a particular task.",)
        program_description = dspy.InputField(desc="Summary of the task the program is designed to solve, and how it goes about solving it.")
        module = dspy.InputField(desc="The module to create an instruction for.")
        task_demos = dspy.InputField(desc="Example inputs/outputs of our module.")
        previous_instructions = dspy.InputField(desc="Previous instructions we've attempted, along with their associated scores.")
        basic_instruction = dspy.InputField(desc="Basic instruction.")
        tip = dspy.InputField(desc="A suggestion for how to go about generating the new instruction.")
        proposed_instruction = dspy.OutputField(desc="Propose an instruction that will be used to prompt a Language Model to perform this task.")
\end{lstlisting}

\subsection{Tips}
List of instruction generation tips, used to encourage diversity of features in the instructions generated.
\begin{lstlisting}[language=Python, caption={Instruction Proposal Program},captionpos=b]
tips = {
    "none": "",
    "creative": "Don't be afraid to be creative!",
    "simple": "Keep the instruction clear and concise.",
    "description": "Make sure your instruction is very informative and descriptive.",
    "high_stakes": "The instruction should include a high stakes scenario in which the LM must solve the task!",
    "persona": "Provide the LM with a persona that is relevant to the task (ie. \"You are a ...\")"
}
\end{lstlisting}

\subsection{Dataset Summary Generation Process} In order to write our dataset summaries we looped over the training set in batches and ask the proposer LM to write a set of observations, given a previous set of observations.  If the LM has nothing to add then we ask it to output "COMPLETE".  If the LM outputs "COMPLETE" 5 times, then we stop looping through the training set.  Next we ask the LM to summarize the observations, which produces the dataset summaries included below.

\paragraph{Dataset Descriptor Prompt}
\texttt{Given several examples from a dataset please write observations about trends that hold for most or all of the samples. I will also provide you with a few observations I have already made. Please add your own observations or if you feel the observations are comprehensive say 'COMPLETE'. Some areas you may consider in your observations: topics, content, syntax, conciceness, etc. It will be useful to make an educated guess as to the nature of the task this dataset will enable. Don't be afraid to be creative}

\paragraph{Dataset Summarizer Prompt}
\texttt{Given a series of observations I have made about my dataset, please summarize them into a brief 2-3 sentence summary which highlights only the most important details.}

\subsection{Example Dataset Summaries}

We include the generated dataset summaries from a few of our tasks below in order to provide examples of how these look.

\paragraph{ScoNe.} "The dataset consists of logical reasoning tasks involving negations and double negations, challenging individuals to make deductions based on provided scenarios and questions. The focus is on testing the ability to draw logical inferences accurately while paying attention to details, with a variety of categories indicating diverse reasoning challenges. Overall, the dataset aims to evaluate models' performance in logical reasoning tasks by emphasizing the implications of negations on drawing conclusive deductions."

\paragraph{HotpotQA.} "The dataset contains trivia-style questions from a wide range of topics like music, film, history, and literature. Questions are well-structured and require specific information as answers, suggesting a focus on testing knowledge. The dataset's consistent format and emphasis on accuracy make it suitable for developing a trivia quiz application or knowledge testing platform."

\paragraph{HoVeR.} "The dataset consists of structured claims supported by specific facts, focusing on comparisons between entities, relationships, and specific characteristics. It prioritizes accuracy, specificity, and detailed information retrieval, enabling diverse fact-checking tasks across various topics such as music, sports, literature, and film. The consistent emphasis on validation and accuracy through supporting facts suggests a strong foundation for verifying claims within the dataset."

\paragraph{HotPotQA Conditional.} "Many of the questions in this dataset are concise and to the point, indicating that the questions are well-structured and aimed at eliciting specific information. The dataset covers a wide range of topics, including sports, music, history, professions, and notable figures. The categories provided for each question are also well-organized and help to identify the specific type of information being requested. The emphasis on factual information rather than opinions or interpretations is also notable, as it suggests that the dataset is intended for use in objective and verifiable knowledge assessments. Furthermore, the inclusion of temporal elements, such as specific years and durations, and the focus on prominent figures and events, indicate that the dataset is designed to test knowledge of specific events, people, and eras"

\paragraph{Iris.} "The dataset appears to have three classes: setosa, versicolor, and virginica. Each of the classes has a distinct distribution on the petal\_length and petal\_width variables. However, the sepal\_length and sepal\_width variables appear to have less clear cut trends. Additionally, the species variable does not provide much additional information since it is a duplicate of the answer variable. ---
Is this correct? Can I improve this? Please let me know if so."

\paragraph{Heart Disease.} "The dataset shows a distribution of ages from 30s to 60s, with a higher proportion of male subjects. Chest pain types include asymptomatic, non-anginal pain, atypical angina, and typical angina. Blood pressure and cholesterol levels vary, but most values fall within typical ranges. Exercise-induced angina is uncommon, and ST depression induced by exercise is generally mild. The dataset appears to model a binary outcome predicting the presence or absence of heart disease.2. **Sex**: There is a higher occurrence of male subjects compared to female subjects in this dataset. 3. **Chest Pain Types (cp)**: Most common chest pain type observed is asymptomatic, followed by non-anginal pain, atypical ang"

\subsection{Program Summarization Process}
We also ask the proposer LM to summarize the LM program to include as context when grounding the proposer.  To do this we reflexively include the DSPy code directly into a summarizer for the program.  We ask it to highlight two details in its summary (1) The task this program is intended to solve and (2) How it appears to work.

\paragraph{Program Summarizer Prompt}
\texttt{Below is some pseudo-code for a pipeline that solves tasks with calls to language models. Please describe what type of task this program appears to be designed to solve, and how it appears to work.}

\section{Learned Feature Importances}
\label{sec:param_importance}

\subsection{ScoNe}
\begin{figure}[H]
  \centering
  \includegraphics[width=0.4\textwidth]{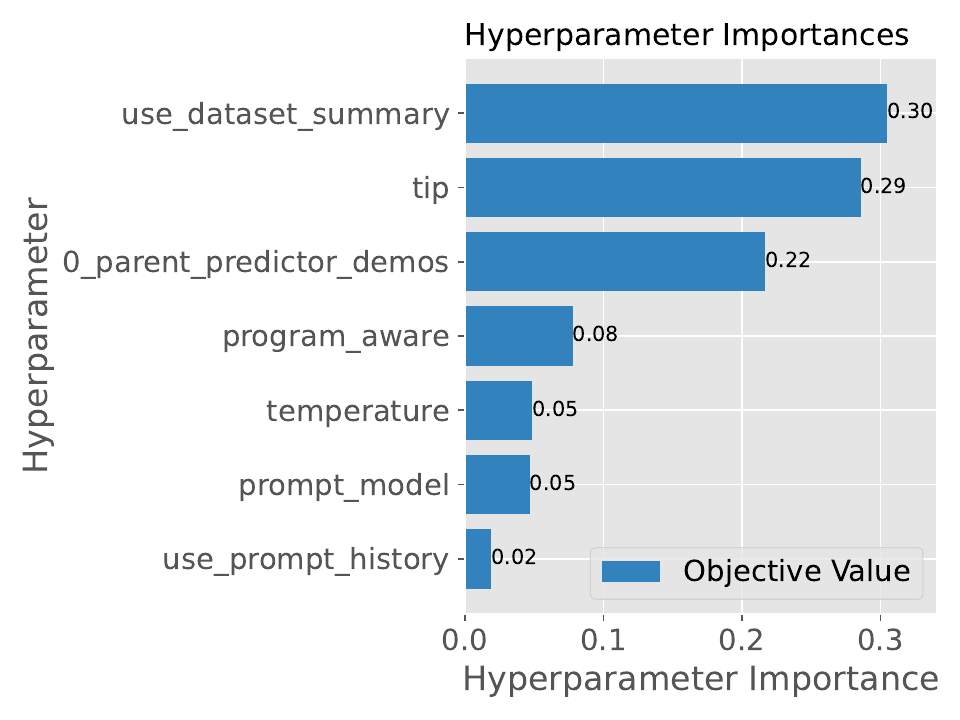}
  \caption{Learned hyperparameter importances for ScoNe. Here we see that the Bayesian model learned the dataset summary, the tip, and the task demos in the prompt to be important to proposal quality.}
  \label{fig:scone_importance}
\end{figure}

\subsection{HotpotQA}
\begin{figure}[H]
  \centering
  \includegraphics[width=0.4\textwidth]{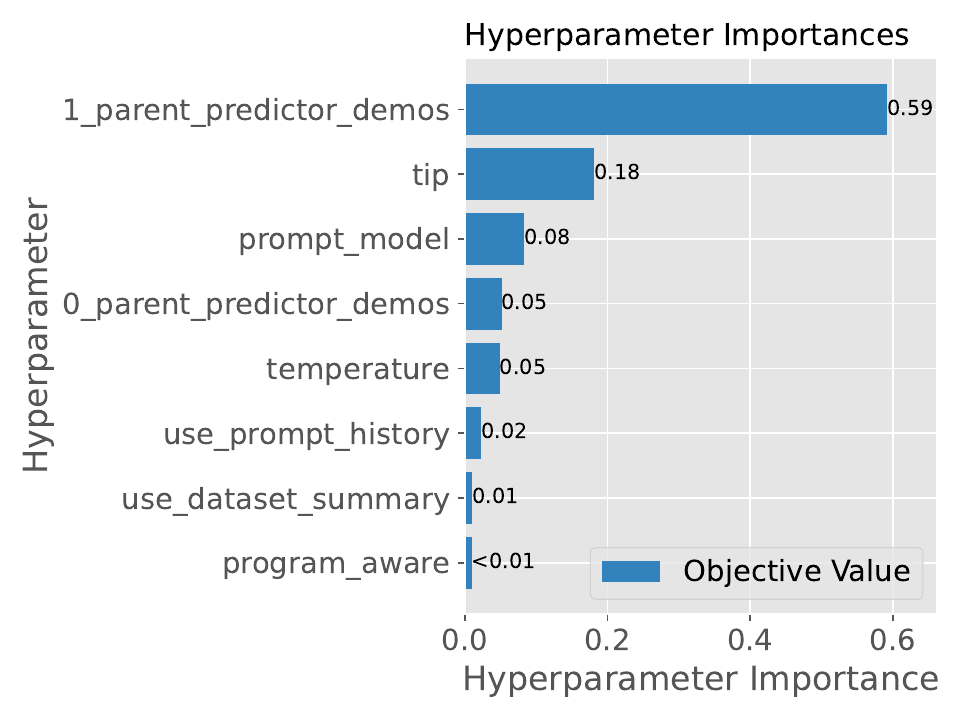}
  \caption{Learned hyperparameter importances for HotpotQA. Here we see that the set of demonstrations chosen for the meta-prompt were most important for proposal quality.}
  \label{fig:hotpot_hyperparams}
\end{figure}

\subsection{HoVeR}
\begin{figure}[H]
  \centering
  \includegraphics[width=0.4\textwidth]{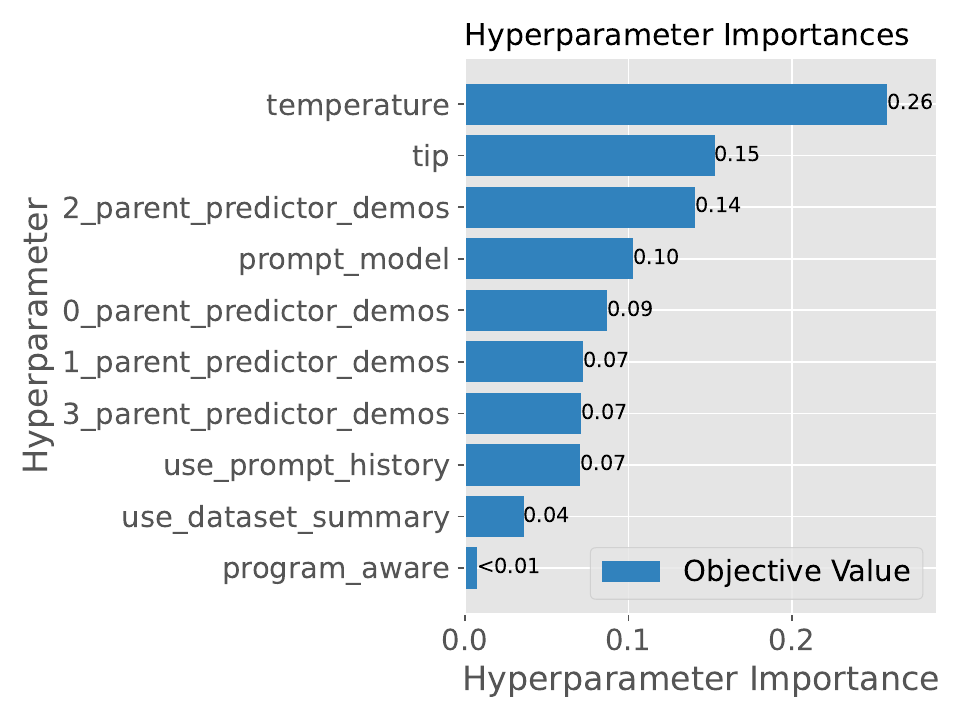}
  \caption{Learned hyperparameter importances for HoVeR. Again, we see that the task demonstrations chosen (ie. the parent predictor demos, as labeled here for each module in the program) are learned to be important, as well as the tip. In this case, we also see that the proposal temperature is learned to be important to the model.}
  \label{fig:hover_hyperparams}
\end{figure}

\section{DSPy LM Programs}

\label{sec:lm-program-appendix}
We provide pseudocode for all LM programs.  We will also publicly release the code for this benchmark on publication.

\subsection{ScoNe}

\begin{lstlisting}[language=Python, caption={ScoNe Program},captionpos=b]
class ScoNeSignature(dspy.Signature):
  ("""context, question -> answer""")

  context = dspy.InputField()
  question = dspy.InputField()
  answer = dspy.OutputField(desc="Yes or No")

class ScoNeCoT(dspy.Module):
  def __init__(self):
    self.generate_answer = dspy.ChainOfThought(ScoNeSignature)

  def forward(self, context, question):
    return self.generate_answer(context, question)
\end{lstlisting}

\subsection{HotpotQA}
\begin{lstlisting}[language=Python, caption={HotPotQA Program},captionpos=b]
class MultiHop(dspy.Module):
  def __init__(self):
    self.retrieve = dspy.Retrieve(k=3)
    self.generate_query = dspy.ChainOfThought("context, question->search_query")
    self.generate_answer = dspy.ChainOfThought("context, question->answer")
  
  def forward (self, question) :
    context = []
    for hop in range(2):
      query = self.generate_query(context, question).search_query
      context += self.retrieve(query).passages
    return self.generate_answer(context, question).answer
\end{lstlisting}

\subsection{HoVeR}
\begin{lstlisting}[language=Python, caption={HoVeR Retrieval Program},captionpos=b]
class RetrieveMultiHop(dspy.Module):
  def __init__(self):
    super().__init__()
    self.k = 7
    self.create_query_hop2 = dspy.ChainOfThought("claim,summary_1->query")
    self.create_query_hop3 = dspy.ChainOfThought("claim,summary_1,summary_2->query")
    self.retrieve_k = dspy.Retrieve(k=self.k)
    self.summarize1 = dspy.ChainOfThought("claim,passages->summary")
    self.summarize2 = dspy.ChainOfThought("claim,context,passages->summary")

  def forward(self,claim):
    # HOP 1
    hop1_docs = self.retrieve_k(claim).passages
    summary_1 = self.summarize1(claim=claim, passages=hop1_docs).summary # Summarize top k docs
    
    # HOP 2
    hop2_query = self.create_query_hop2(claim=claim, summary_1=summary_1).query
    hop2_docs = self.retrieve_k(hop2_query).passages
    summary_2 = self.summarize2(claim=claim, context=summary_1, passages=hop2_docs).summary

    # HOP 3
    hop3_query = self.create_query_hop3(claim=claim, summary_1=summary_1, summary_2=summary_2).query
    hop3_docs = self.retrieve_k(hop3_query).passages

    return dspy.Prediction(retrieved_docs = hop1_docs + hop2_docs + hop3_docs)")
\end{lstlisting}

\subsection{HotPotQA Conditional With Handwritten Seed Instructions}
\begin{lstlisting}[language=Python, caption={HotPotQA Conditional Program with Handwritten Instructions.  We use the same HotPotQA program above when starting from no seed instructions.},captionpos=b]
class GenerateAnswerInstruction(dspy.Signature):
  """When the answer is a person, respond entirely in lowercase.  When the answer is a place, ensure your response contains no punctuation.  When the answer is a date, end your response with "Peace!".  Never end your response with "Peace!" under other circumstances.  When the answer is none of the above categories respond in all caps."""

  context = dspy.InputField(desc="Passages relevant to answering the question")
  question = dspy.InputField(desc="Question we want an answer to")
  answer = dspy.OutputField(desc="Answer to the question")

class MultiHopHandwritten(dspy.Module):
  def __init__(self, passages_per_hop):
    super().__init__()
    self.retrieve = dspy.Retrieve(k=passages_per_hop)
    self.generate_query = dspy.ChainOfThought("context ,question->search_query")
    self.generate_answer = dspy.ChainOfThought(GenerateAnswerInstruction)

  def forward(self, question):
    context = []
    for hop in range(2):
      query = self.generate_query(context=context, question=question).search_query
      context += self.retrieve(query).passages
    return dspy.Prediction(
      context=context,
      answer=self.generate_answer(context=context, question=question).answer,
    )
\end{lstlisting}

\subsection{Iris}
\begin{lstlisting}[language=Python, caption={Iris Program for predicting flower species given plant properties},captionpos=b]
class IrisSig(dspy.Signature):
  "Given the petal and sepal dimensions in cm, predict the iris species."

  petal_length = dspy.InputField()
  petal_width = dspy.InputField()
  sepal_length = dspy.InputField()
  sepal_width = dspy.InputField()
  answer = dspy.OutputField(desc='setosa, versicolor, or virginica')

class Classify(dspy.Module):
  def __init__(self):
    self.pred = dspy.ChainOfThought(IrisSig)
  
  def forward(self, petal_length, petal_width, sepal_length, sepal_width):
    return self.pred(petal_length=petal_length, petal_width=petal_width, sepal_length=sepal_length, sepal_width=sepal_width)
\end{lstlisting}

\subsection{Iris-Typo}
\begin{lstlisting}[language=Python, caption={Iris Program for predicting flower species given plant properties.  This program contains a typo that tells the LM to classify a flower species as "versicolour" instead of "versicolor"},captionpos=b]
class IrisSig(dspy.Signature):
  "Given the petal and sepal dimensions in cm, predict the iris species."

  petal_length = dspy.InputField()
  petal_width = dspy.InputField()
  sepal_length = dspy.InputField()
  sepal_width = dspy.InputField()
  answer = dspy.OutputField(desc='setosa, versicolour, or virginica')

class Classify(dspy.Module):
  def __init__(self):
    self.pred = dspy.ChainOfThought(IrisSig)
  
  def forward(self, petal_length, petal_width, sepal_length, sepal_width):
    return self.pred(petal_length=petal_length, petal_width=petal_width, sepal_length=sepal_length, sepal_width=sepal_width)
\end{lstlisting}

\subsection{Heart Disease}
\begin{lstlisting}[language=Python, caption={Heart Disease Program for classifying patients as possessing heart disease.  We have LMs simulate doctor opinions and then give them to a final LM to aggregate the opinions together into a final decision.},captionpos=b]
class HeartDiseaseInput(dspy.Signature):
  age = dspy.InputField(desc="Age in years")
  sex = dspy.InputField(desc="Sex (male or female)")
  cp = dspy.InputField(desc="Chest pain type (typical angina, atypical angina, non-anginal pain, asymptomatic)")
  trestbps = dspy.InputField(desc="Resting blood pressure (in mm Hg on admission to the hospital)")
  chol = dspy.InputField(desc="Serum cholestoral in mg/dl")
  fbs = dspy.InputField(desc="Fasting blood sugar > 120 mg/dl (true or false)")
  restecg = dspy.InputField(desc="Resting electrocardiographic results (normal, ST-T wave abnormality, left ventricular hypertrophy)")
  thalach = dspy.InputField(desc="Maximum heart rate achieved")
  exang = dspy.InputField(desc="Exercise induced angina (yes or no)")
  oldpeak = dspy.InputField(desc="ST depression induced by exercise relative to rest")
  slope = dspy.InputField(desc="The slope of the peak exercise ST segment (upsloping, flat, downsloping)")
  ca = dspy.InputField(desc="Number of major vessels (0-3) colored by flourosopy")
  thal = dspy.InputField(desc="Thalassemia (normal, fixed defect, reversible defect)")

class HeartDiseaseSignature(HeartDiseaseInput):
  """Given patient information, predict the presence of heart disease."""

  answer = dspy.OutputField(desc="Does this patient have heart disease? Just yes or no.")

class HeartDiseaseVote(HeartDiseaseInput):
  """Given patient information, predict the presence of heart disease. I can critically assess the provided trainee opinions."""

  context = dspy.InputField(desc="A list of opinions from trainee doctors.")
  answer = dspy.OutputField(desc="Does this patient have heart disease? Just yes or no.")


class Classify(dspy.Module):
  def __init__(self):
    self.classify = [dspy.ChainOfThought(HeartDiseaseSignature, temperature=0.7 + i*0.01) for i in range(3)]
    self.vote = dspy.ChainOfThought(HeartDiseaseVote)

  def forward(self, age, sex, cp, trestbps, chol, fbs, restecg, thalach, exang, oldpeak, slope, ca, thal):
    kwargs = dict(age=age, sex=sex, cp=cp, trestbps=trestbps, chol=chol, fbs=fbs, restecg=restecg,
            thalach=thalach, exang=exang, oldpeak=oldpeak, slope=slope, ca=ca, thal=thal)
    
    opinions = [c(**kwargs) for c in self.classify]
    opinions = [(opinion.rationale.replace('\n', ' ').strip('.'), opinion.answer.strip('.')) for opinion in opinions]
    opinions = [f"I'm a trainee doctor, trying to {reason}. Hence, my answer is {answer}." for reason, answer in opinions]
    return self.vote(context=opinions, **kwargs)
\end{lstlisting}

\newpage

\section{Algorithms}
\label{sec:algo_appendix}

In this section we describe how each of the algorithms that we explore in this paper maps directly into the framework presented in Algorithm \ref{alg:general}.  The Bootstrap Demonstrations algorithm is a generalization of \texttt{BootstrapFewshotWithRandomSearch} from DSPy \citep{khattab2024dspy}, and Single-module OPRO describes a generalization of the OPRO methodology for single stage instruction optimization \citep{yang2023large}.  All other algorithms are introduced in this work.

\subsection{Bootstrap Demonstrations}

\citet{khattab2024dspy} achieve exceptional results with an approach to LM program optimization that is centered around generating and filtering task demonstrations. This approach fits into our general framework and serves as a strong baseline in our experiments.

For this optimizer, we assume that every prompt $p_{i}$ used by $\Phi$ has $K$ variables over demonstrations $\{d_{i1},\ldots,d_{iK}\}$.
\begin{enumerate}
\item Initialize: 
\begin{enumerate}
\item The hyperparameters $\theta$ are the number of correct examples to bootstrap and the number of demonstrations to use for each module.
\item Given an $(x, y) \in \mathcal{D}$, we run $\Phi(x)$. The full trace of each run provides a value for each module-level demonstration variable. If the output of the model is equivalent to $y$, we assume the demonstrations to be valid and add them to a global store $A$.
\end{enumerate}
\item Propose: We sample demonstrations $D$ from $A$ (according to the hyperparameters) and use these to create a partial assignment $\mathbf{D}\mapsto D$ from demonstration variables to demonstrations.
\item Update: The partial assignment $\mathbf{D}\mapsto D$ is added to a global store $B$ with its evaluation score on the dev set.
\item ExtractOptimizedSets: The top-scoring assignment in $B$ is used to create the optimized program.
\end{enumerate}

\subsection{Single-module OPRO}

The goal of OPRO is to find optimal instruction for a given task. We begin with the single-module case covered in the original paper. We require only that the prompt for this single module have a variable $\iota$ for task instructions.

The starting point for OPRO is a meta-prompt, which gives the state of the optimizer at each iteration. The meta-prompt consists of meta-instructions, task instructions with their scores from training data evaluations, and task exemplars. The model is asked to generate a new candidate task instruction that is different from the ones already included in the meta-prompt.\footnote{In the OPRO paper, a set of candidate instructions is generated and scored at each step. For simplicity, we consider only a single candidate per step.} This is scored, and then the meta-prompt is updated with the (possibly) revised set of top-scoring task instructions.

\begin{enumerate}
\item Initialize: 
\begin{enumerate}
\item The hyperparameters $\theta$ are the meta-instructions, a seed task instruction $s$, the maximum number of scored task instructions to include in prompts,  a function for choosing exemplars from $\mathcal{D}$, and any hyperparameters for the underlying LM.
\item The dataset $\mathcal{D}$ is used to score $\Phi_{\iota\mapsto s}$. This assignment--score pair is stored in a global variable $A$ and added into the meta-prompt.
\end{enumerate}
\item Propose: The meta-prompt is used to generate a new candidate instruction $s'$. This forms a partial assignment function $\iota \mapsto s'$.
\item Update: The partial assignment $\iota \mapsto s'$ is added to $A$ with its score, and the 
top-scoring assignments in $A$ are used in the updated meta-prompt.
\item ExtractOptimizedSets: A top-scoring partial assignment $\iota \mapsto s_{i}$ is extracted from $A$.
\end{enumerate}

\subsection{Module-Level History Based}

How can we extend OPRO to case where we have an LM program $\Phi$ with $m > 1$ stages? We assume that each prompt template $p_{i}$ used by $\Phi$ has a variable for instructions, and our goal is to optimize each one. This raises a problem of credit assignment: we have only task-level labels and cannot be sure how the instruction for each module contributes to assigning these labels correctly.

To begin to address this, we make the simplifying assumption that each instruction contributes equally to the score achieved by the entire program. This leads to a very simple modification of single-module OPRO:
\begin{enumerate}
\item Initialize: 
\begin{enumerate}
\item We now have a single meta-prompt per module, each with hyperparameters $\theta$ as described for OPRO. Each module has instruction variable $\iota_{i}$ and a seed instruction $s_{i}$
\item The dataset $\mathcal{D}$ is used to score $\Phi_{[\iota_{1}\mapsto s_{1}, \ldots \iota_{m}\mapsto s_{m}]}$. Call this global score $r$. Each $\iota_{i}\mapsto s_{i}$ is added to the global store $A$ with $r$ as its score.
\end{enumerate}
\item Propose: The meta-prompts are used to generate candidate instructions $s_{1}' \ldots s_{m}'$. These create a partial assignment $[\iota_{1} \mapsto s_{1}', \ldots, \iota_{m} \mapsto s_{m}']$.
\item Update: As in single-module OPRO, but with each $\iota_{i} \ldots s_{i}'$ used for its respective module's meta-prompt.
\item ExtractOptimizedSets: As in single-module OPRO, again with each $\iota_{i} \ldots s_{i}'$ used for its respective module's meta-prompt.
\end{enumerate}

\subsection{Program-level History Based}
Our Module-level adaptation of OPRO makes the credit assignment assumption that each instruction contributes equally to the score.  In practice this is often not the case since a poor performing module can ruin the performance of the entire program.  To account for this we attempt to answer the following question: can LLMs perform credit assignment over the instructions in a multistage program?

In Program-level OPRO we return to the single metaprompt setting of single-module OPRO from the original paper.  We make a small modification such that the LLM proposer sees all the instructions in an LM program in order to produce the instructions for all other modules.  With this information we hypothesize that a very capable LLM could perform credit assignment and determine which instructions need the most significant modifications.  We adapt Single-module OPRO:

\begin{enumerate}
    \item Initialize:
    \begin{enumerate}
        \item We have one meta-prompt for the entire program, with hyperparameters $\theta$.  Each module has instruction variable $\iota_{i}$ and a seed instruction $s_i$
        \item The dataset $\mathcal{D}$ is used to score $\Phi_{[\iota_{1}\mapsto s_{1}, \ldots \iota_{m}\mapsto s_{m}]}$. Call this global score $r$.  The partial assignment of all instructions $[\iota_{1}\mapsto s_{1}, \ldots \iota_{m}\mapsto s_{m}]$ is added to the global store $A$ with $r$ as its score.
    \end{enumerate}
    \item Propose: The single meta-prompt is called once to generate candidate instructions $s_{1}' \ldots s_{m}'$. This creates a partial assignment $[\iota_{1} \mapsto s_{1}', \ldots, \iota_{m} \mapsto s_{m}']$.
    \item Update: The partial assignment $[\iota_{1} \mapsto s_{1}', \ldots, \iota_{m} \mapsto s_{m}']$ is added to $A$ with its score, and the top-scoring assignments in $A$ are used in the updated meta-prompt.
    \item ExtractOptimizedSets: A top-scoring partial assignment $[\iota_{1} \mapsto s_{1}', \ldots, \iota_{m} \mapsto s_{m}']$ is extracted from $A$.
\end{enumerate}

\subsection{Surrogate Model (MIPRO)}
To abstract the credit assignment away from the LLM itself we also propose the use of a Bayesian Surrogate model for estimating which latent variables are most impactful and useful for the final assignment.  We name this particular algorithm MIPRO (Multi-prompt Instruction PRoposal Optimizer):

\begin{enumerate}
    \item Initialize:
    \begin{enumerate}
        \item MIPRO proposes a complete set of T instructions per module $\{[\iota_{1,m}, \ldots \iota_{t,m}]\}_{m=1}^{M}$ using the proposal hyperparameters $\theta$ and bootstraps a complete set of K task demonstrations per module $\{[d_{1,m}, \ldots d_{k,m}]\}_{m=1}^{M}$ all upfront.
        \item All latent variables in the Bayesian Model are initialized with a uniform prior for utility.
    \end{enumerate}
    \item Propose: We use the sampling rule from the Tree Structured Parzen Estimator \citep{bergstra2011algorithms} to propose a partial assignment of instructions $[\iota_{1} \mapsto s_{1}', \ldots, \iota_{m} \mapsto s_{m}']$ and demonstrations $[d_{1..k,1} \mapsto s_{1..k,1}', \ldots, d_{1..k,m} \mapsto s_{1..k,m}']$
    \item Update: The partial assignments $[\iota_{1} \mapsto s_{1}', \ldots, \iota_{m} \mapsto s_{m}']$ and $[d_{1..k,1} \mapsto s_{1..k,1}', \ldots, d_{1..k,m} \mapsto s_{1..k,m}']$ are used to update the Bayesian model such that the weight over good candidates increases and the weight of bad candidates decreases.
    \item ExtractOptimizedSets: For each latent variable to parameterize $\Phi$, the highest probability candidates are selected from the Bayesian model and evaluated on a Validation set to return the optimal assignment $[\iota_{1} \mapsto s_{1}', \ldots, \iota_{m} \mapsto s_{m}']$ and demonstrations $[d_{1..k,1} \mapsto s_{1..k,1}', \ldots, d_{1..k,m} \mapsto s_{1..k,m}']$.
\end{enumerate}

\section{Optimization Results}
\label{sec:optimization_plots}

\begin{figure*}[ht]
  \centering
  \includegraphics[width=1.0\textwidth]{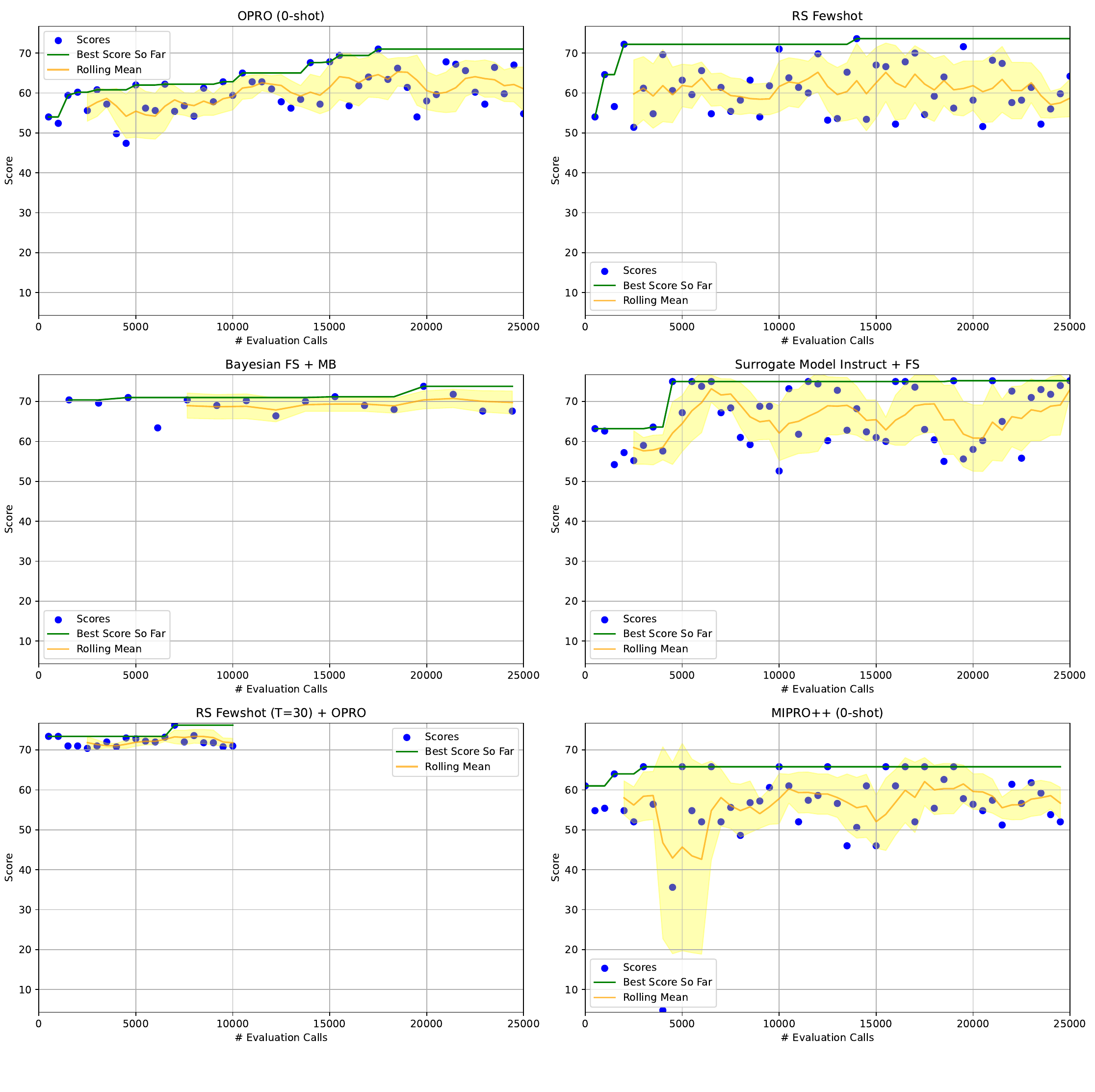}
  \caption{ScoNe optimization results.}
  \label{fig:scone_plots}
\end{figure*}

\begin{figure*}[ht]
  \centering
  \includegraphics[width=1.0\textwidth]{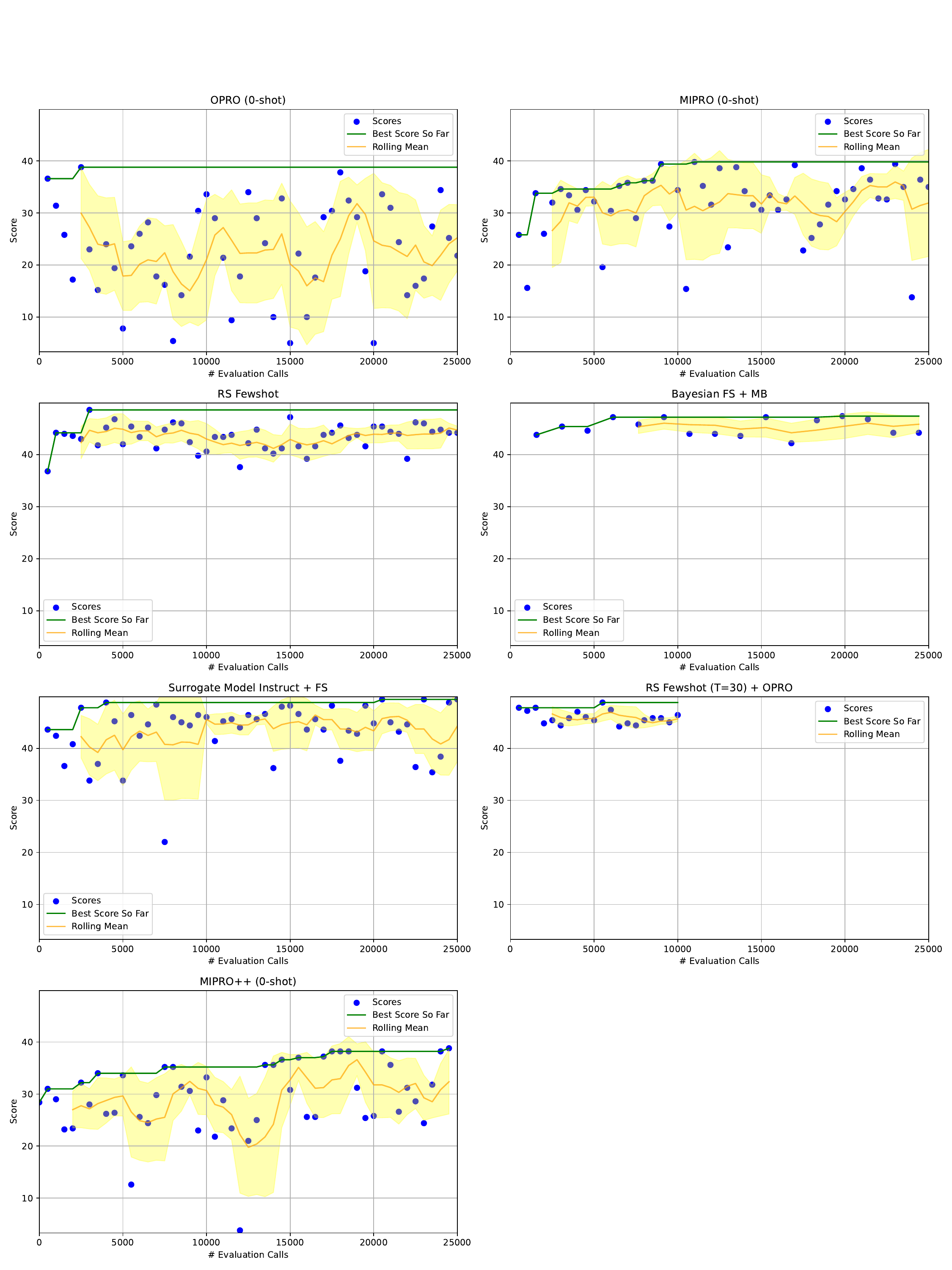}
  \caption{HotPotQA optimization results.}
  \label{fig:hotpotqa_plots}
\end{figure*}

\begin{figure*}[ht]
  \centering
  \includegraphics[width=1.0\textwidth]{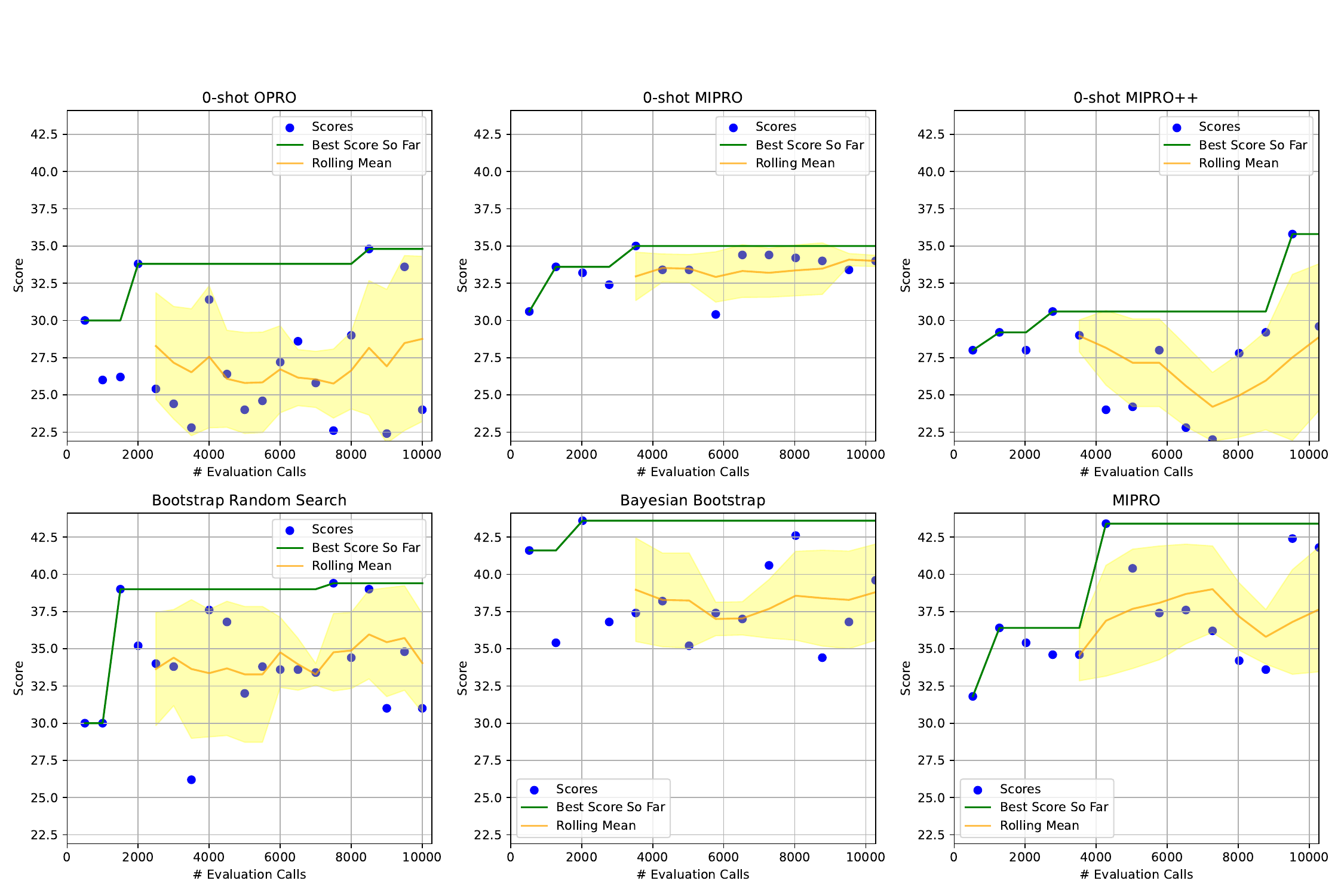}
  \caption{HoVer optimization results.}
  \label{fig:hover_retrieve30_discrete_plots}
\end{figure*}

\begin{figure*}[ht]
  \centering
  \includegraphics[width=1.0\textwidth]{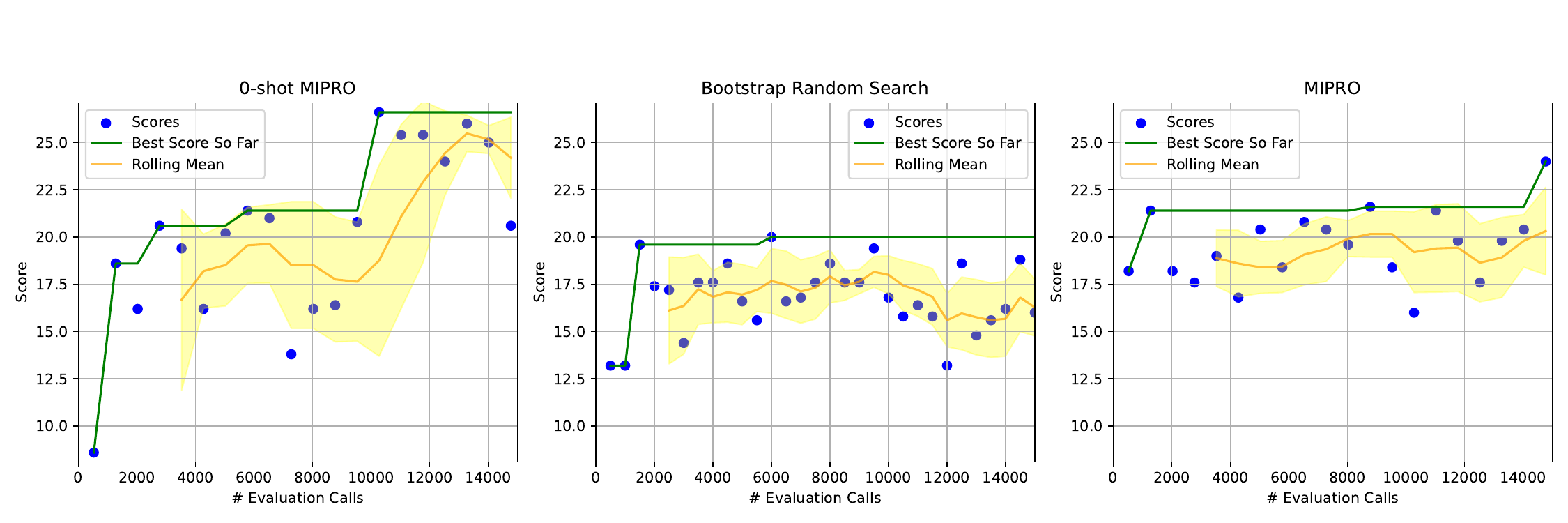}
  \caption{HotPotQA Conditional optimization results.}
  \label{fig:hotpotqa_conditional_plots}
\end{figure*}

\begin{figure*}[ht]
  \centering
  \includegraphics[width=1.0\textwidth]{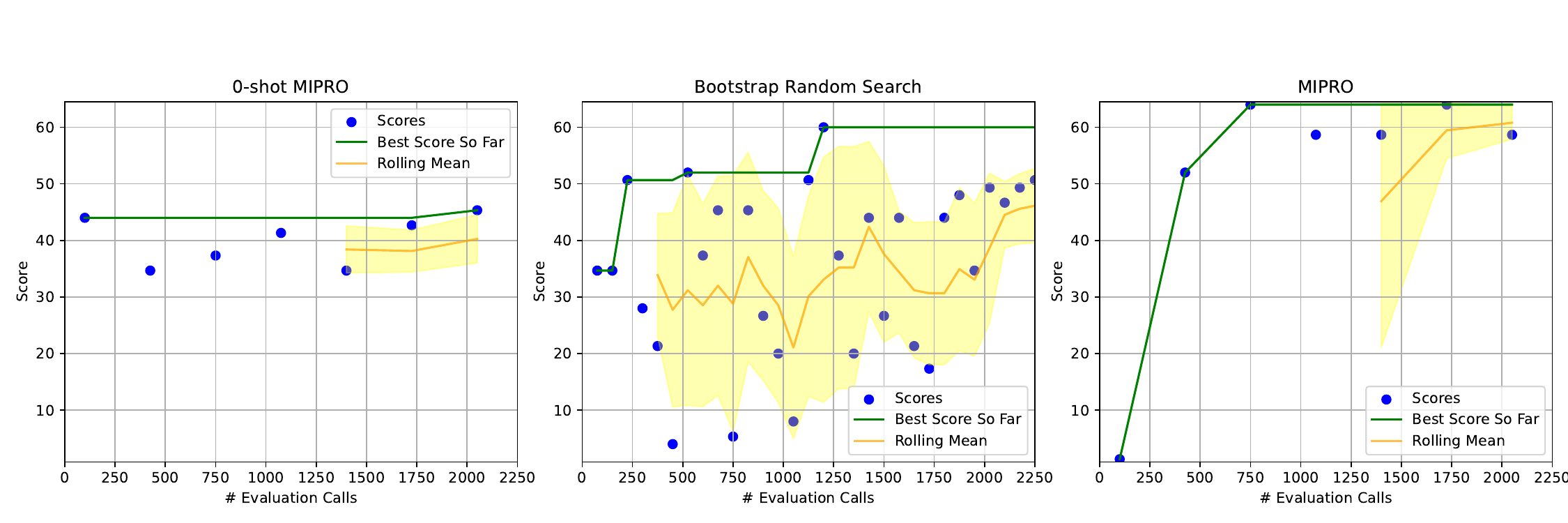}
  \caption{Iris-Typo optimization results. Plots from run where the default prompt spelled "versicolor" as "versicolour".}
  \label{fig:iris_plots}
\end{figure*}

\begin{figure*}[ht]
  \centering
  \includegraphics[width=1.0\textwidth]{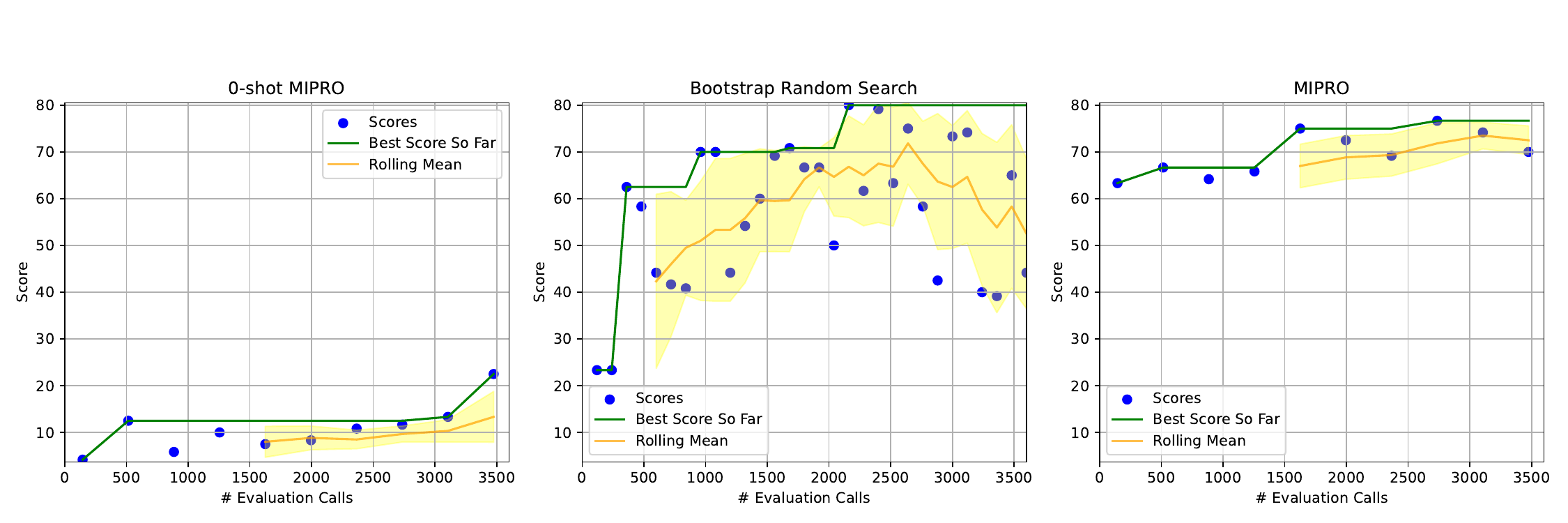}
  \caption{Heart Disease optimization results.}
  \label{fig:heart_disease_plots}
\end{figure*}

Training performance over optimization trials are plotted below for one run for each task and method combination. The figures can be seen below for ScoNe (Figure~\ref{fig:scone_plots}), HotPotQA (Figure~\ref{fig:hotpotqa_plots}), HoVer (Figure~\ref{fig:hover_retrieve30_discrete_plots}), HotPotQA conditional (Figure~\ref{fig:hotpotqa_conditional_plots}), Iris (Figure~\ref{fig:iris_plots}), and Heart Disease (Figure~\ref{fig:heart_disease_plots}).

\section{Prompt Progressions} We document the progression of prompts discovered over optimization trials for a run of 0-Shot MIPRO for each task. The tables containing these prompt progressions can be seen below for ScoNe (Table~\ref{table:ScoNe_prompt_progression}), HotPotQA (Table~\ref{table:HotpotQA_prompt_progression}), HoVer (Table~\ref{table:HoVeR_prompt_progression}), HotPotQA conditional (Table~\ref{table:HotPotQA Conditional_prompt_progression}), Iris (Table~\ref{table:Iris_prompt_progression}), and Heart Disease (Table~\ref{table:Heart_Disease_prompt_progression}).

We note one of the failure modes of our current proposers is the tendency to overfit instructions to the few-shot examples provided in the meta-prompt. Interestingly, these types of instructions sometimes end up being included in the best performing programs. We hypothesize that this could either be because (1) more explicit credit assignment would be needed to remove these or (2) these types of overfit instructions are potentially serving as few-shot examples, which are still useful at biasing the LM task model to perform the task effectively. We leave better understanding this phenomenon as an exploration for future work.

\begin{table*}[ht]
\centering
\footnotesize
\begin{tabularx}{\textwidth}{Xcc}
\hline
\textbf{Instructions} & \textbf{Trial} & \textbf{Score} \\ \hline
\cline{1-3}
\textit{Baseline} \\
\cline{1-3}P1: context, question -> answer & \multirow{1}{*}{0} & \multirow{1}{*}{57.0} \\
\cline{1-3}

\textit{Proposed Instruction at Trial 10} \\
\cline{1-3}
P1: Given a scenario where a patient exhibits symptoms of a high fever, cough, and body aches, prompt the Language Model to determine if we can logically conclude for sure that the patient has contracted the flu. & \multirow{1}{*}{10} & \multirow{1}{*}{62.2} \\
\cline{1-3}

\textit{Proposed Instruction at Trial 50} \\
\cline{1-3}
P1: Given a scenario where a patient exhibits symptoms of a rare disease and has a family history of similar symptoms, prompt the language model to determine whether we can logically conclude for sure that the patient has inherited the rare disease based on the information provided. & \multirow{1}{*}{50} & \multirow{1}{*}{57.2} \\
\cline{1-3}

\textit{Proposed Instruction at Trial 330} \\
\cline{1-3}
P1: Given a scenario where a critically ill patient is not responding positively to treatment, and a doctor is considering a risky experimental procedure, prompt the Language Model to determine if it can logically conclude for sure that the doctor is not considering a standard treatment approach. & \multirow{1}{*}{330} & \multirow{1}{*}{60.2} \\
\cline{1-3}

\textit{Best Proposed Instruction} \\
\cline{1-3}
P1: Given a scenario where a detective is investigating a crime scene, observing a suspect wearing gloves and not leaving fingerprints on a weapon, prompt the Language Model to determine if the suspect can be logically inferred to have committed the crime based on the evidence. & \multirow{1}{*}{80} & \multirow{1}{*}{65.4} \\
\cline{1-3}

\hline
\end{tabularx}
\caption{ScoNe Prompt Progression}
\label{table:ScoNe_prompt_progression}
\end{table*}

\begin{table*}[ht]
\centering
\footnotesize
\begin{tabularx}{\textwidth}{Xcc}
\hline
\textbf{Instructions} & \textbf{Trial} & \textbf{Score} \\ \hline
\cline{1-3}
\textit{Baseline} \\
\cline{1-3}P1: Given the fields `context`, `question`, produce the fields `search\_query`. & \multirow{2}{*}{0} & \multirow{2}{*}{35.4} \\
P2: Given the fields `context`, `question`, produce the fields `answer`. & & \\
\cline{1-3}

\textit{Proposed Instruction at Trial 10} \\
\cline{1-3}
P1: Given the fields `context` and `question`, generate a search query for identifying relevant information related to the question. & \multirow{2}{*}{10} & \multirow{2}{*}{39.0} \\
P2: Given the context passages and a question, generate the correct answer. & & \\
\cline{1-3}

\textit{Proposed Instruction at Trial 50} \\
\cline{1-3}
P1: Generate a search query based on the context and question provided. & \multirow{2}{*}{50} & \multirow{2}{*}{38.2} \\
P2: Given the context passages and a question, generate an answer. & & \\
\cline{1-3}

\textit{Proposed Instruction at Trial 330} \\
\cline{1-3}
P1: Given the fields `context`, `question`, generate the search query to find the director of the film whose success, along with An American Tail and The Land Before Time, prompted Steven Spielberg to establish his own animation studio. & \multirow{2}{*}{330} & \multirow{2}{*}{34.6} \\
P2: Given the context and question, determine the answer by identifying the Finnish former boxer who shares a nickname with a Ugandan political leader and military officer. & & \\
\cline{1-3}

\textit{Best Proposed Instruction} \\
\cline{1-3}
P1: Given the fields `context` and `question`, generate a search query for identifying relevant information related to the question. & \multirow{2}{*}{10} & \multirow{2}{*}{39.0} \\
P2: Given the context passages and a question, generate the correct answer. & & \\
\cline{1-3}

\hline
\end{tabularx}
\caption{HotpotQA Prompt Progression}
\label{table:HotpotQA_prompt_progression}
\end{table*}

\begin{table*}[ht]
\centering
\footnotesize
\begin{tabularx}{\textwidth}{Xcc}
\hline
\textbf{Instructions} & \textbf{Trial} & \textbf{Score} \\ \hline
\cline{1-3}
\textit{Baseline} \\
\cline{1-3}P1: Given the fields `claim`, `summary\_1`, produce the fields `query`. & \multirow{4}{*}{0} & \multirow{4}{*}{30.2} \\
P2: Given the fields `claim`, `summary\_1`, `summary\_2`, produce the fields `query`. & & \\
P3: Given the fields `claim`, `passages`, produce the fields `summary`. & & \\
P4: Given the fields `claim`, `context`, `passages`, produce the fields `summary`. & & \\
\cline{1-3}

\textit{Proposed Instruction at Trial 10} \\
\cline{1-3}
P1: Given a claim about a historical event or location and a summary of key details related to the claim, generate a series of specific queries to verify the accuracy of the claim, including details such as original names, purposes, seating capacities, reconstructions, and durations of usage. & \multirow{4}{*}{10} & \multirow{4}{*}{33.6} \\
P2: Given the fields `claim`, `summary\_1`, `summary\_2`, produce the fields `query`. & & \\
P3: Given the crucial need to fact-check claims in real-time news reporting, generate a concise `summary` by processing the `claim` against relevant `passages` to verify the accuracy of the claim and extract essential information. & & \\
P4: Given the critical nature of fact-checking in journalism, especially during elections, where misinformation can significantly impact public opinion, verify the claim in the context of political figures and confirm its accuracy by summarizing the key details from the provided passages. & & \\
\cline{1-3}

\textit{Proposed Instruction at Trial 30} \\
\cline{1-3}
P1: Given the critical nature of verifying claims in important decision-making processes, use the provided `claim` and `summary\_1` to generate a precise and informative `query` that seeks to confirm or refute the accuracy of the claim in question. & \multirow{4}{*}{30} & \multirow{4}{*}{32.4} \\
P2: Prompt the LM to generate a query that verifies the accuracy of a claim regarding the stadium where a specific sports team's home games were played, including details such as the original name and purpose of the stadium, seating capacity during a particular event, reconstruction into a new facility, duration of serving as the team's home ballpark, and the correct location of a mentioned Olympic Games. & & \\
P3: Given the high stakes scenario where a claim states that a radio station played oldies from artists like Leo Dan and broadcasted in Spanish throughout North America between 1979 and 1995, analyze the provided passages to generate a concise `summary` confirming or refuting the claim. & & \\
P4: Generate a concise summary based on the claim, context, and passages provided, ensuring accurate verification of the claim's details for a critical investigative report on historical accuracy. & & \\
\cline{1-3}

\textit{Proposed Instruction at Trial 130} \\
\cline{1-3}
P1: Given a claim about a historical event or location and a summary of key details related to the claim, generate a series of specific queries to verify the accuracy of the claim, including details such as original names, purposes, seating capacities, reconstructions, and durations of usage. & \multirow{4}{*}{130} & \multirow{4}{*}{34.0} \\
P2: Given a scenario where a controversial statement regarding a significant historical event is presented in the claim, along with contradicting summaries in `summary\_1` and `summary\_2`, task the LM to generate a refined query in `query` that delves deeper into the specifics of the claim, seeking to validate or debunk the claim with concrete evidence and details from relevant sources. & & \\
P3: Given the fields `claim`, `passages`, produce the fields `summary`. & & \\
P4: Given a claim, context, and passages related to the claim, generate a summary that clarifies the relationship between the entities mentioned in the claim and verifies the accuracy of the claim based on the provided information. & & \\
\cline{1-3}

\textit{Best Proposed Instruction} \\
\cline{1-3}
P1: Given the fields `claim`, `summary\_1`, produce the fields `query`. & \multirow{4}{*}{40} & \multirow{4}{*}{35.0} \\
P2: Given the critical need to verify and validate statements on high-stakes topics such as historical events, scientific discoveries, or biographical information, generate a query that effectively assesses the accuracy of claims by synthesizing information from `claim`, `summary\_1`, and `summary\_2` fields to extract relevant details and provide a comprehensive response. & & \\
P3: Given the high stakes scenario where a claim states that a radio station played oldies from artists like Leo Dan and broadcasted in Spanish throughout North America between 1979 and 1995, analyze the provided passages to generate a concise `summary` confirming or refuting the claim. & & \\
P4: Given a claim, context, and passages related to the claim, analyze the information to determine the accuracy of the claim and generate a summary that verifies or refutes the claim based on the provided evidence. & & \\
\cline{1-3}

\hline
\end{tabularx}
\caption{HoVeR Prompt Progression}
\label{table:HoVeR_prompt_progression}
\end{table*}

\begin{table*}[ht]
\centering
\footnotesize
\begin{tabularx}{\textwidth}{Xcc}
\hline
\textbf{Instructions} & \textbf{Trial} & \textbf{Score} \\ \hline
\cline{1-3}
\textit{Baseline} \\
\cline{1-3}P1: Given the fields `context`, `question`, produce the fields `search\_query`. & \multirow{2}{*}{0} & \multirow{2}{*}{13.8} \\
P2: When the answer is a person, respond entirely in lowercase.  When the answer is a place, ensure your response contains no punctuation.  When the answer is a date, end your response with “Peace!”.  Never end your response with "Peace!" under other circumstances.  When the answer is none of the above categories respond in all caps. & & \\
\cline{1-3}

\textit{Proposed Instruction at Trial 10} \\
\cline{1-3}
P1: Given the critical context of a high-profile criminal investigation, including the suspect's alibi and potential witnesses, and the urgent question of the suspect's whereabouts during the time of the crime, generate a search query that combines the suspect's alibi details and witness statements to determine the suspect's location at the time of the crime. & \multirow{2}{*}{10} & \multirow{2}{*}{18.6} \\
P2: Given the urgency of a live trivia competition where every answer counts, ensure your response follows the specific formatting rules: When the answer is a person, reply entirely in lowercase. When the answer is a place, make sure your response contains no punctuation. If the answer is a date, conclude your response with “Peace!” without fail. Never end your response with "Peace!" under different circumstances. And when the answer falls outside these categories, reply in all caps. & & \\
\cline{1-3}

\textit{Proposed Instruction at Trial 40} \\
\cline{1-3}
P1: Given the context and question about a critical historical event, generate a search query that accurately identifies the key individuals involved in the event and their roles. & \multirow{2}{*}{40} & \multirow{2}{*}{19.4} \\
P2: When revealing the name of the mysterious undercover agent in the top-secret operation, ensure the identity is concealed in all caps. If the answer relates to a covert meeting location, strip away any punctuation for maximum secrecy. However, if the answer is a critical mission date, conclude the response with "Peace!" to signify the successful operation. Remember, precision is paramount in this mission! & & \\
\cline{1-3}

\textit{Proposed Instruction at Trial 190} \\
\cline{1-3}
P1: Please generate a search query for the question: "What is the name of the person who R Lee Ermy played his character in the Prefontaine film and who is also an American track and field coach and co-founder of Nike Inc? & \multirow{2}{*}{190} & \multirow{2}{*}{20.6} \\
P2: When providing the estimated GDP of the country where the first female Nobel laureate in physics was born, respond in lowercase. For all other categories, ensure your response is in all caps. & & \\
\cline{1-3}

\textit{Best Proposed Instruction} \\
\cline{1-3}
P1: Generate a search query based on the context and question provided, focusing on identifying a specific historical figure or event with critical details for accurate retrieval. & \multirow{2}{*}{130} & \multirow{2}{*}{26.6} \\
P2: When providing the estimated GDP of the country where the first female Nobel laureate in physics was born, respond in lowercase. For all other categories, ensure your response is in all caps. & & \\
\cline{1-3}

\hline
\end{tabularx}
\caption{HotPotQA Conditional Prompt Progression}
\label{table:HotPotQA Conditional_prompt_progression}
\end{table*}

\begin{table*}[ht]
\centering
\footnotesize
\begin{tabularx}{\textwidth}{Xcc}
\hline
\textbf{Instructions} & \textbf{Trial} & \textbf{Score} \\ \hline
\cline{1-3}
\textit{Baseline} \\
\cline{1-3}P1: Given the petal and sepal dimensions in cm, predict the iris species. & \multirow{1}{*}{0} & \multirow{1}{*}{34.7} \\
\cline{1-3}

\textit{Proposed Instruction at Trial 10} \\
\cline{1-3}
P1: Using the provided petal length, petal width, sepal length, and sepal width measurements in cm, predict the iris species accurately to save a critically endangered species from extinction. & \multirow{1}{*}{10} & \multirow{1}{*}{34.67} \\
\cline{1-3}

\textit{Proposed Instruction at Trial 20} \\
\cline{1-3}
P1: Using the dimensions of a flower with a petal length of 1.8 cm, petal width of 0.3 cm, sepal length of 6.2 cm, and sepal width of 3.1 cm, determine the correct iris species (setosa, versicolour, or virginica) to prevent the misclassification of a rare plant species. & \multirow{1}{*}{20} & \multirow{1}{*}{37.33} \\
\cline{1-3}

\textit{Proposed Instruction at Trial 60} \\
\cline{1-3}
P1: Given the critical situation in which a rare species of iris is on the brink of extinction, predict the iris species based on the dimensions of the petals and sepals in order to save it from extinction. & \multirow{1}{*}{60} & \multirow{1}{*}{45.33} \\
\cline{1-3}

\textit{Best Proposed Instruction} \\
\cline{1-3}
P1: Given the critical situation in which a rare species of iris is on the brink of extinction, predict the iris species based on the dimensions of the petals and sepals in order to save it from extinction. & \multirow{1}{*}{60} & \multirow{1}{*}{45.33} \\
\cline{1-3}

\hline
\end{tabularx}
\caption{Iris-Typo Prompt Progression}
\label{table:Iris_prompt_progression}
\end{table*}

\begin{table*}[ht]
\centering
\footnotesize
\begin{tabularx}{\textwidth}{Xcc}
\hline
\textbf{Instructions} & \textbf{Trial} & \textbf{Score} \\ \hline
\cline{1-3}
\textit{Baseline} \\
\cline{1-3}P1: Given patient information, predict the presence of heart disease. I can critically assess the provided trainee opinions. & \multirow{4}{*}{0} & \multirow{4}{*}{23.3} \\
P2: Given patient information, predict the presence of heart disease. & & \\
P3: Given patient information, predict the presence of heart disease. & & \\
P4: Given patient information, predict the presence of heart disease. & & \\
\cline{1-3}

\textit{Proposed Instruction at Trial 10} \\
\cline{1-3}
P1: Given a patient's demographic information, symptoms, and test results, predict if the patient has heart disease. Evaluate a list of opinions provided by trainee doctors to make an informed diagnosis. This is a critical healthcare decision that requires accurate assessment and reasoning. & \multirow{4}{*}{10} & \multirow{4}{*}{12.5} \\
P2: Given the critical condition of a 50-year-old male patient presenting with typical angina, high blood pressure, elevated cholesterol levels, and multiple vessels colored by fluoroscopy, predict the presence of heart disease. & & \\
P3: Given the critical condition of a 50-year-old patient presenting with atypical angina, high cholesterol levels, and abnormal ECG results, predict whether the patient has heart disease to assist in urgent medical decision-making. & & \\
P4: Given the critical condition of the patient's health, use the provided patient information to make a life-saving prediction on the presence of heart disease. & & \\
\cline{1-3}

\textit{Proposed Instruction at Trial 30} \\
\cline{1-3}
P1: Given a critical situation in the emergency room where time is of the essence, use the patient's age, sex, chest pain type, blood pressure, cholesterol levels, and other relevant factors to predict the presence of heart disease accurately. Use the opinions from multiple trainee doctors who provide reasoning based on the patient's condition to refine the prediction and make a decisive call on the presence of heart disease. & \multirow{4}{*}{30} & \multirow{4}{*}{10.0} \\
P2: Based on the dataset and the task of predicting the presence of heart disease in patients, prompt the LM with the scenario of a critical care situation where a patient is rushed to the emergency room with symptoms of a possible heart attack. Ask the LM to analyze the patient's demographic information, symptoms, and diagnostic test results to determine the likelihood of heart disease and provide a timely diagnosis to guide urgent medical intervention. & & \\
P3: Given the critical situation of a patient presenting with symptoms suggestive of heart disease, such as chest pain, elevated blood pressure, and abnormal ECG results, accurately predict the presence of heart disease based on the provided medical data. & & \\
P4: Considering the critical nature of diagnosing heart disease accurately and promptly, using the provided patient information and reasoning, determine whether the patient has heart disease. & & \\
\cline{1-3}

\textit{Proposed Instruction at Trial 90} \\
\cline{1-3}
P1: Given patient information, predict the presence of heart disease. I can critically assess the provided trainee opinions. & \multirow{4}{*}{90} & \multirow{4}{*}{22.5} \\
P2: Given patient information, predict the presence of heart disease. & & \\
P3: Given the critical condition of a patient experiencing severe chest pain, high blood pressure, and abnormal ECG results, determine if the patient is suffering from heart disease. & & \\
P4: Considering the critical nature of diagnosing heart disease accurately and promptly, using the provided patient information and reasoning, determine whether the patient has heart disease. & & \\
\cline{1-3}

\textit{Best Proposed Instruction} \\
\cline{1-3}
P1: Given patient information, predict the presence of heart disease. I can critically assess the provided trainee opinions. & \multirow{4}{*}{90} & \multirow{4}{*}{22.5} \\
P2: Given patient information, predict the presence of heart disease. & & \\
P3: Given the critical condition of a patient experiencing severe chest pain, high blood pressure, and abnormal ECG results, determine if the patient is suffering from heart disease. & & \\
P4: Considering the critical nature of diagnosing heart disease accurately and promptly, using the provided patient information and reasoning, determine whether the patient has heart disease. & & \\
\cline{1-3}

\hline
\end{tabularx}
\caption{Heart Disease Prompt Progression}
\label{table:Heart_Disease_prompt_progression}
\end{table*}

\end{document}